\documentclass[fleqn,10pt]{wlscirep}
\usepackage[english]{babel}
\usepackage[utf8]{inputenc}
\usepackage[T1]{fontenc}
\usepackage{hyperref}
\usepackage{array}
\usepackage{booktabs} 
\usepackage{hyperref} 
\usepackage{longtable}
\usepackage{amssymb} 
\usepackage{pifont}    
\usepackage{xcolor}    
\usepackage{doi}
\usepackage[detect-all]{siunitx}
\usepackage{comment}
\usepackage{multirow}
\usepackage{makecell}
\usepackage[symbol]{footmisc}
\usepackage{makecell}

\usepackage{etoolbox}
\newtoggle{arxiv}
\toggletrue{arxiv}
\newcommand{\flexiref}[2]{\iftoggle{arxiv}{\autoref{#1}}{section \textit{#2}}}
\newcommand{\appref}[1]{\hyperref[#1]{Appendix~\ref*{#1}}}
\newcommand{\flexisuppref}[2]{\iftoggle{arxiv}{\appref{#1}}{supplementary information#2}}
\newcommand{\mysection}[1]{\iftoggle{arxiv}{\section{#1}}{\section*{#1}}}
\newcommand{\mysubsection}[1]{\iftoggle{arxiv}{\subsection{#1}}{\subsection*{#1}}}
\newcommand{\mysubsubsection}[1]{\iftoggle{arxiv}{\subsubsection{#1}}{\subsubsection*{#1}}}
\newcommand{\citep}[1]{\cite{#1}}
\newcommand{\citet}[1]{\cite{#1}}

\newcounter{supsection}
\newcounter{supsubsection}[supsection]
\renewcommand{\thesupsection}{S\arabic{supsection}}
\renewcommand{\thesupsubsection}{S\arabic{supsection}.\arabic{supsubsection}}
\newcounter{supsubsubsection}[supsubsection]
\renewcommand{\thesupsubsubsection}{S\arabic{supsection}.\arabic{supsubsection}.\arabic{supsubsubsection}}
\newcounter{suppfigure}[supsection]
\newcounter{supptable}[supsection]

\newcommand{\supsection}[1]{%
  \refstepcounter{supsection}%
  \setcounter{suppfigure}{1}%
  \setcounter{supptable}{1}%
  \section*{\thesupsection\quad #1}%
  \addcontentsline{toc}{section}{\thesupsection\quad #1}%
}

\newcommand{\supsubsection}[1]{%
  \refstepcounter{supsubsection}%
  \subsection*{\thesupsubsection\quad #1}%
  \addcontentsline{toc}{subsection}{\thesupsubsection\quad #1}%
}

\newcommand{\supsubsubsection}[1]{%
  \refstepcounter{supsubsubsection}%
  \subsubsection*{\thesupsubsubsection\quad #1}%
  \addcontentsline{toc}{subsubsection}{\thesupsubsubsection\quad #1}%
}

\definecolor{myblue}{HTML}{0072B2}

\DeclareUnicodeCharacter{00394}{$\Delta$}

\title{A multi-modal dataset for insect biodiversity with imagery and DNA at the trap and individual level}

\author[1,*]{Johanna~Orsholm}
\author[2,3,*]{John~Quinto}
\author[4]{Hannu~Autto}
\author[1]{Gaia~Banelyte}
\author[1]{Nicolas~Chazot}
\author[5,6]{Jeremy~deWaard}
\author[5,7]{Stephanie~deWaard}
\author[1]{Arielle~Farrell}
\author[8]{Brendan~Furneaux}
\author[9]{Bess~Hardwick}
\author[5]{Nao~Ito}
\author[3,10,11]{Amlan~Kar}
\author[4]{Oula~Kalttopää}
\author[1]{Deirdre~Kerdraon}
\author[12]{Erik~Kristensen}
\author[5]{Jaclyn~McKeown}
\author[9]{Tommi~Mononen}
\author[1]{Ellen~Nein}
\author[1]{Hanna~Rogers}
\author[1,9]{Tomas~Roslin}
\author[1]{Paula~Schmitz}
\author[5]{Jayme~Sones}
\author[4]{Maija~Sujala}
\author[5]{Amy~Thompson}
\author[5]{Evgeny~V.~Zakharov}
\author[2,3]{Iuliia~Zarubiieva}
\author[2,3]{Akshita~Gupta}
\author[3]{Scott~C.~Lowe}
\author[2,3]{Graham~W.~Taylor}

\affil[1]{Department of Ecology, Swedish University of Agricultural Sciences, Uppsala, Sweden}
\affil[2]{University of Guelph, Guelph, Ontario, Canada}
\affil[3]{Vector Institute, Toronto, Ontario, Canada}
\affil[4]{Kilpisjärvi Biological Station, University of Helsinki, Helsinki, Finland}
\affil[5]{Centre for Biodiversity Genomics, University of Guelph, Guelph, Ontario, Canada}
\affil[6]{Current affiliation: Department of Entomology, National Museum of Natural History, Smithsonian Institution, United States}
\affil[7]{Current affiliation: University of Guelph, Guelph, Ontario, Canada}
\affil[8]{Department of Biological and Environmental Science, University of Jyväskylä, Jyväskylä, Finland}
\affil[9]{Faculty of Biological and Environmental Sciences, University of Helsinki, Finland}
\affil[10]{NVIDIA, Toronto, Ontario, Canada}
\affil[11]{University of Toronto, Toronto, Ontario, Canada}
\affil[12]{Unit for Field-based Forest Research, Swedish University of Agricultural Sciences, Umeå, Sweden}
\affil[*]{These authors contributed equally}

\begin{abstract} 
Insects comprise millions of species, many experiencing severe population declines under environmental and habitat changes. High-throughput approaches are crucial for accelerating our understanding of insect diversity, with DNA barcoding and high-resolution imaging showing strong potential for automatic taxonomic classification. However, most image-based approaches rely on individual specimen data, unlike the unsorted bulk samples collected in large-scale ecological surveys. We present the Mixed Arthropod Sample Segmentation and Identification (MassID45) dataset for training automatic classifiers of bulk insect samples. It uniquely combines molecular and imaging data at both the unsorted sample level and the full set of individual specimens. Human annotators, supported by an AI-assisted tool, performed two tasks on bulk images: creating segmentation masks around each individual arthropod and assigning taxonomic labels to over \num{17000} specimens. Combining the taxonomic resolution of DNA barcodes with precise abundance estimates of bulk images holds great potential for rapid, large-scale characterization of insect communities. This dataset pushes the boundaries of tiny object detection and instance segmentation, fostering innovation in both ecological and machine learning research.
\end{abstract}
\begin{document}

\pagestyle{plain}

\flushbottom
\maketitle

\thispagestyle{empty}

\renewcommand{\sectionautorefname}{Section}
\let\subsectionautorefname\sectionautorefname
\let\subsubsectionautorefname\sectionautorefname

\mysection{Background} 

Insects are the most diverse organism group on Earth, with more than one million described species \cite{COL} and an estimated four million species yet to be discovered \cite{stork2018}. Climate change and anthropogenic activities are causing rapid declines in insect populations \cite{Wagner2020}, and many species likely face extinction in the coming decades \cite{cardoso2020}. Yet our understanding of insect diversity is seriously data-limited and incomplete. In particular, data generation is hampered by a severe lack of available taxonomic expertise \cite{quentin2004, pearson2011}. Consequently, there is an urgent need for high-throughput methods to study and monitor insect communities, with machine learning-based image classification and molecular methods quickly advancing as promising tools for the taxonomic characterization of samples. 

To taxonomically classify diverse insect samples, substantial amounts of training data are needed. This requires taxonomically labelled examples --- images for visual classification and DNA sequences for molecular approaches. At present, the most abundant insect DNA sequence data consists of short fragments (typically 658 bp) from a standardized gene region, the cytochrome c oxidase subunit 1 (COI), commonly known as DNA barcodes \cite{Hebert2003}. Comprehensive datasets containing millions of images and DNA barcodes from individual insect specimens are now available for training classifiers \cite{Gharaee2024bioscan1m, gharaee2024bioscan5mmultimodaldatasetinsect, Steinke2024.07.15.600863}. However, many large-scale ecological studies or monitoring efforts collect insects in bulk samples, which contain multiple specimens with mixed taxonomic composition. Sorting bulk samples into individual specimens requires substantial effort. Thus, for many ecological applications, there is currently a discrepancy between the available specimen-level training data and the bulk-level samples in need of classification. To bridge this gap, taxonomically annotated bulk-level training images are needed.

Capturing images of unsorted bulk samples is a straightforward procedure, yielding a single image depicting all specimens from the sample, hereafter referred to as a \emph{bulk image}. Classifying insects from such bulk images presents two key challenges. First, given insects' small size and the high specimen density in bulk samples, individual specimens appear as tiny objects within the image, with restricted morphological detail. Second, insects' high taxonomic diversity requires large training datasets to achieve adequate coverage of different taxa by the classifier. Together, these pose a significant challenge for the taxonomic annotation of images. As a consequence, previous studies attempting insect classification from bulk images have worked with a substantially reduced number of classes \cite{schneider2022,Fujisawa2023}, resulting in more coarse information than what is often needed for community ecology. 

DNA barcode data can also be generated from bulk samples, a process called DNA metabarcoding \cite{riaz2011}. The resulting DNA barcodes can then be classified into taxa by comparison with extensive reference databases, such as the Barcode of Life Data Systems (BOLD) \cite{Ratnasingham2024}. Assuming that the target group is well-represented in the reference data, DNA metabarcoding produces detailed information about the taxonomic composition of the sample. It also yields data on the relative abundances of different sequence variants. However, one of the main drawbacks of DNA metabarcoding is that it is challenging to infer absolute abundances across samples and taxa \cite{Luo2023}. To address this challenge, we propose to integrate DNA and image data, leveraging the strength of both modalities. Combining detailed taxonomic classifications from DNA metabarcoding with bulk images of the same samples, where counting of taxa is relatively straightforward, could enable absolute, taxon-specific abundance estimates. Using both images and barcodes has been shown to enhance classifier performance compared to using either data source in isolation \cite{Badirli2023ClassifyingTU,Gong2025-ml}. Further, when using individual-level data of both modalities, the genus of unknown species was predicted with more than 80\% success \cite{Badirli2023ClassifyingTU}. This could allow researchers to explore the taxonomic composition at higher ranks even in samples which contain a large number of unknown species, thus enabling studies of particularly diverse ecosystems and organism groups. An additional benefit of combining molecular and image data is the potential to extract trait data, such as the body size distribution across individuals in a sample \cite{schneider2022}. 

Here, we present the Mixed Arthropod Sample Segmentation and Identification (MassID45) dataset, centred on 45 bulk arthropod samples (primarily insects) collected using Malaise traps deployed in Sweden and Finland in 2021. Harnessing both imaging and molecular information, each sample features DNA metabarcoding data and one or more unsorted bulk images. We further provide sample-level biomass measurements, which may support trait-based analyses. To facilitate the training of machine learning-based classifiers, we additionally supply individual-level images and DNA barcode sequences obtained after sorting each sample into individual specimens (\num{36402} in total). Leveraging AI-assisted annotation and sample-specific DNA-based classifications, we provide a detailed segmentation mask and taxonomic annotation for each arthropod in the bulk images. Using the bulk-level data, we benchmark instance segmentation on tiny objects, drawing on related methods for object detection. Our dataset and experiments provide a valuable perspective on the domain-specific challenges of capturing tiny, densely arranged, and sometimes overlapping objects --- an area less emphasized in standard object detection datasets. Recent advances in tiny object detection \cite{tong2022, tresson2021}, open-world detection \cite{Gupta2022}, multi-modal learning \cite{Gong2025-ml, Badirli2023ClassifyingTU}, and AI-assisted annotation \cite{Kirillov2023} collectively allow researchers to better detect, classify, and analyze numerous minuscule specimens in a single frame. This dataset is poised to support a wide range of ecological applications, such as training automated classifiers for bulk samples, accurately counting specimens in large collections, and enabling large-scale morphological analyses. From a machine learning standpoint, it also opens up avenues in developing instance segmentation methods for tiny objects, exploiting weakly labelled information for classification, and open-world detection using fine-grained classes. By bridging these fields, our resource provides a valuable foundation for further research in both ecology and machine learning.

\mysection{Methods}

\begin{figure*}[tb!]
	\centering    
	\includegraphics[width=1\textwidth]{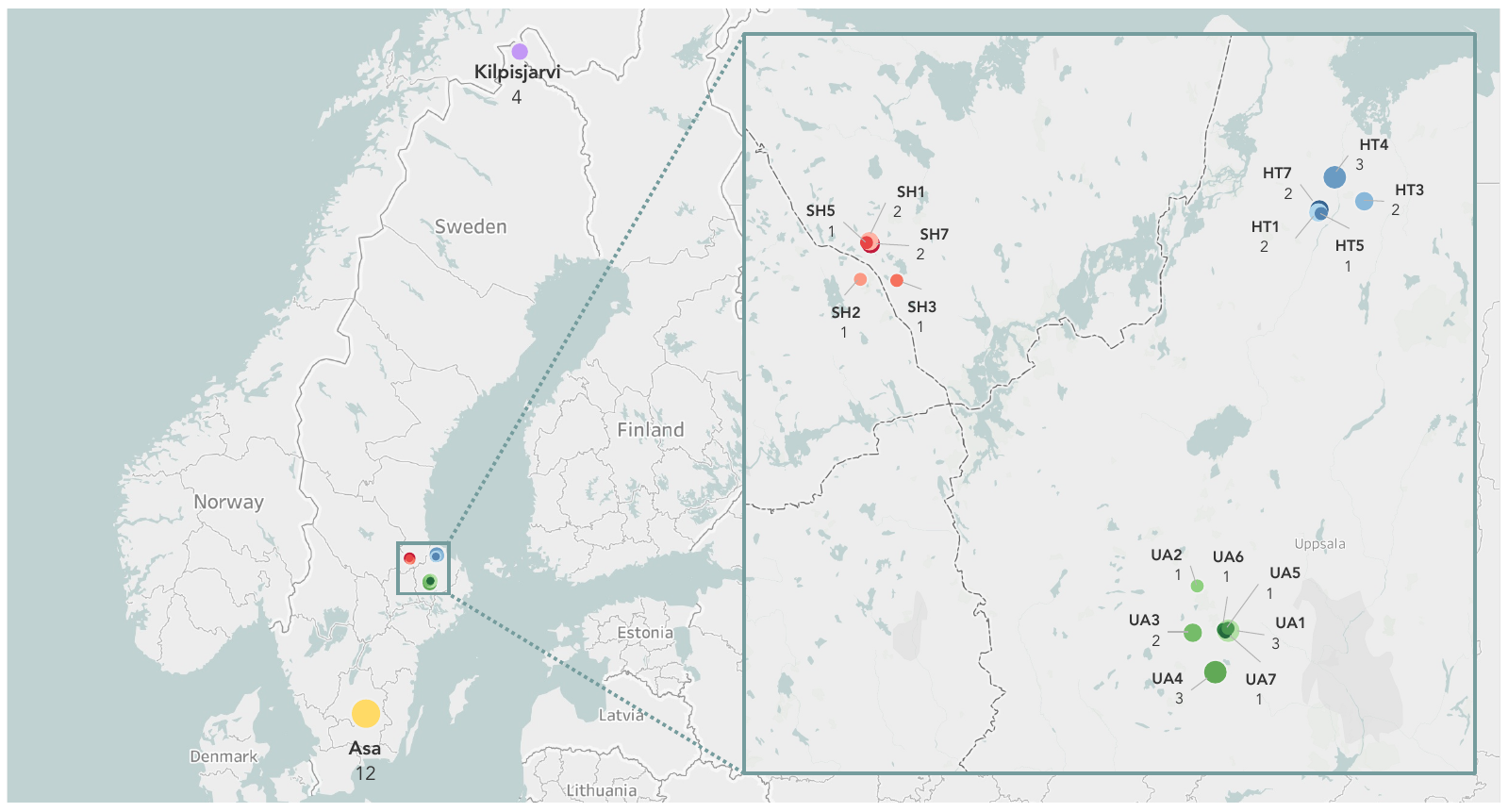}
 \includegraphics[width=1\textwidth]{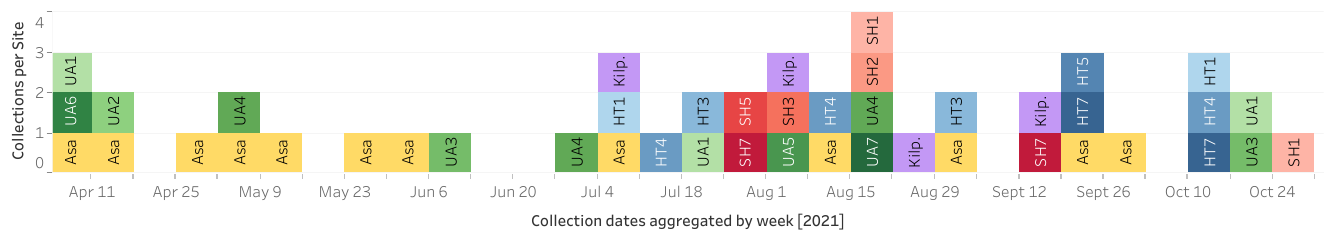}\caption{Geographical distribution and collection dates of samples. Top: Locations of the 19 sampling sites across Sweden and northern Finland. SH, HT, and UA are part of a hierarchical sampling design, each including 5--7 trap locations. The size of each circle is proportional to the number of samples collected at that site, which is also indicated by an integer below the trap name.
 Bottom: Temporal distribution of the MassID45 samples. Collection dates have been aggregated by week so that samples collected during the same week are displayed in the same column, regardless of what day of the week they were collected. Hierarchically organized sites (SH, HT, and UA) are coloured with shades of the same main colours to emphasize their geographical proximity.}
 \label{fig:map}
\end{figure*}

\mysubsection{Spatiotemporal sampling information}
We sampled arthropod communities at 19 sites in Sweden and northern Finland using Townes-style Malaise traps \cite{banelyte2023}. We deployed the traps continuously during 2021 and emptied them once per week. The MassID45 dataset we present here constitutes a subset of 45 of these samples, collected between March 31 and October 25, 2021 (Figure~\ref{fig:map}). Each sample is uniquely named with a six-character alphanumeric code and has associated geographic and temporal information, including the latitude and longitude of the sampling site, as well as placement and collection dates. 17 of the 19 sampling sites were part of a hierarchical sampling design \cite{hardwick2025} and were therefore clustered relatively closely together. In ecological analyses, this nested design allows us to compare samples which are close to each other (some $10^2$ m) to sites which are farther away ($10^6$ m) through a range of scales (Figure~\ref{fig:map}). After collection, samples were shipped to the Centre for Biodiversity Genomics, Guelph, Canada, where they were preserved in fresh 96\% ethanol and stored at --20°C until analysis. 

\mysubsection{DNA barcoding and imaging workflows}

\begin{figure*}[tb!]
	\centering    
 \includegraphics[width=.95\textwidth]{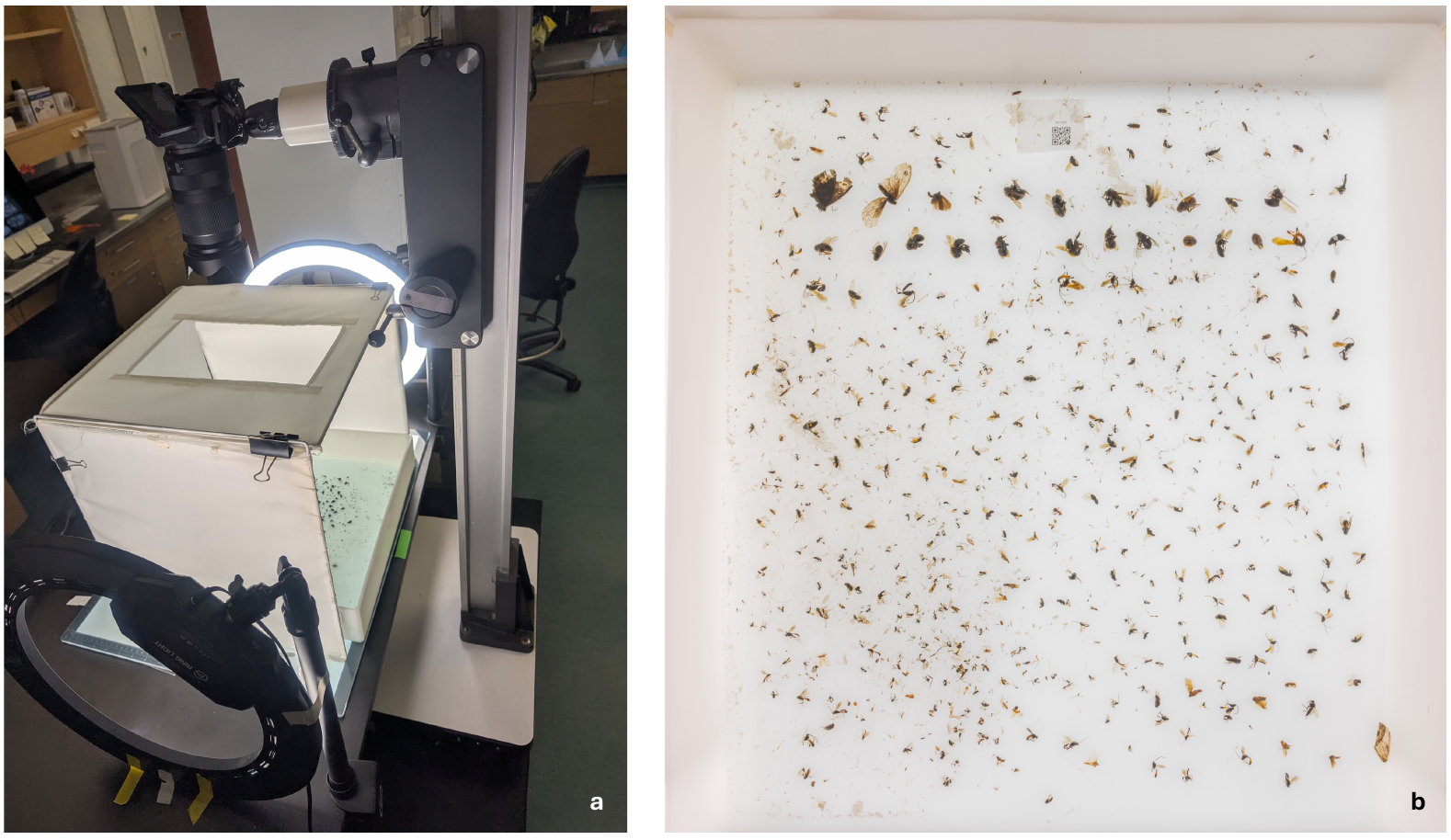}
 \caption{(\textbf{a}) Imaging setup used to capture bulk images of the MassID45 dataset, including the positioning of the camera, light cube and ring light sources. (\textbf{b}) A representative image captured using the described imaging setup, with the sides trimmed.}
 \label{fig:setup}
\end{figure*}

\mysubsubsection{Bulk sample analysis}
We first analyzed samples through a bulk workflow \cite{dewaard2024}, where all specimens collected in a sample were analyzed simultaneously without prior sorting. We weighed the arthropods from the bulk sample after filtering out the ethanol to obtain the wet biomass. We then performed non-destructive lysis for DNA extraction and collected three technical replicates from each sample. After extraction, we amplified a short (418 bp) fragment within the standard barcoding region of COI, which we then sequenced on an Illumina NovaSeq 6000. After DNA extraction, we transferred each sample to a translucent sorting tray (44 $\times$ 39 cm) with a shallow layer of ethanol and carefully spread out the specimens to minimize overlap. We placed the tray on an LED panel inside a modified light cube, where the front panel was removed and a hole was added in the ceiling to fit a camera  (Figure~\ref{fig:setup}a). To further improve light conditions, we used two ring lights placed on opposing sides of the light cube. We captured a bulk image from above with a Canon EOS R5 camera and an RF 24--240 mm F4-6.3 IS USM zoom lens mounted on a large copy stand. We used the following camera settings: focal length of 27 mm, aperture f/20, shutter speed 1/6 seconds, and ISO 100. Each photo included a QR code unique to the sample (Figure~\ref{fig:setup}b). For four samples weighing more than approximately 10~g, we divided the sample into two sorting trays, resulting in two bulk images for each of these samples, for a total of 49 images.

We manually edited the full-resolution RAW images (45 megapixels; $8192\times5464$) in Adobe Lightroom Classic to improve contrast and ensure visibility of both light and dark insect body parts, using the following settings: we increased exposure by 1.3 stops, set whites and highlights to \num{-100}, and shadows to \num{+50}. To restore image contrast and colour we also adjusted clarity and saturation to 20 and increased the white balance from 4200K to 5050K. To reduce noise and purple fringing, we applied luminance noise reduction and defringe values of 20. Finally, we increased sharpening to 60 and saved the images in JPEG format. 

\mysubsubsection{Individual specimen analysis}\label{sec:inddata}
After bulk imaging was completed, we placed each specimen from the bulk samples in a separate well in a 96-well microplate for individual analysis. Specimens smaller than 5 mm were placed directly in the well and imaged using a Keyence VHX-7000 Digital Microscope system \cite{Steinke2024.07.15.600863}. For larger specimens (approximately $>$5 mm), we removed a single leg for DNA extraction and pinned the main body of the arthropod for imaging using an automatic Imaging Rig \cite{Steinke2024ImagingRig}. We amplified and sequenced full 658-bp DNA barcodes for each specimen using single-molecule real-time (SMRT) sequencing \cite{hebert2018} on a PacBio Sequel platform. The success rate of amplification and sequencing of DNA barcodes from the individual specimens was 97.5\%, though only 89.6\% passed quality and contamination checks. In combination with factors affecting metabarcoding success, such as primer bias and amplification of non-target DNA (e.g., gut contents), we therefore expect some discrepancies between the individual and bulk-level DNA barcoding data. We uploaded images and DNA barcodes to BOLD and assigned taxonomic classifications based on both image and molecular information using the BOLD ID engine. We retained all specimens for future morphological reference in the natural history collection of the Centre for Biodiversity Genomics (BIOUG).

\textit{Sample-specific taxonomies.} Using individual-level DNA barcodes, we constructed sample-specific taxonomies to guide the annotations of the corresponding bulk images. First, we compiled a base taxonomy containing the ranks kingdom, phylum, subphylum, class, order, suborder, infraorder, superfamily, family, subfamily, genus, and species, starting with the taxonomy provided by BOLD \cite{Ratnasingham2024}. We then supplemented the taxonomy with the ranks suborder, infraorder and superfamily from Dyntaxa, the Swedish taxonomy database (downloaded from \url{https://artfakta.se/}), which covers all arthropods recorded from Sweden. We subset the Dyntaxa taxonomy to Hexapoda and Arachnida and combined it with the BOLD taxonomy by matching genus names within phyla and classes. Any taxonomic discordances between the taxonomies were resolved by giving the BOLD taxonomy precedence as follows. We used family names as they occurred in the BOLD taxonomy, and manually checked by comparison with the NCBI taxonomy database \cite{schoch2020ncbi} that the suborder, infraorder, and superfamily from Dyntaxa were correct for all cases where Dyntaxa used a different family name. If NCBI listed another suborder, infraorder, or superfamily for the BOLD family name, we changed the discordant rank in our taxonomy to match NCBI. However, we did not add any information from NCBI to ranks that were empty in our taxonomy. If there was no taxonomic information for the BOLD family, we kept the information from Dyntaxa for suborder, infraorder, and superfamily. Diplura, Collembola, and Protura occurred as classes in BOLD but as orders in Dyntaxa. We kept them as classes in our taxonomy and removed all sub- and infraorders, as the same taxa appeared as orders in BOLD. The orders Phthiraptera and Psocoptera in the Dyntaxa taxonomy were combined into order Psocodea in BOLD. We therefore used the latter in our taxonomy. We also added ``microlepidoptera'' as an informal taxonomic group between the ranks of infraorder and superfamily. This group included 14 superfamilies within the order Lepidoptera (Adeloidea, Choreutoidea, Gelechioidea, Gracillarioidea, Micropterigoidea, Nepticuloidea, Pterophoroidea, Pyraloidea, Schreckensteinioidea, Tineoidea, Tischerioidea, Tortricoidea, Urodoidea, and Yponomeutoidea). While microlepidoptera is not a true taxonomic group, it is a useful classification when working with insect images with low resolution. Finally we generated sample-specific taxonomies by using the taxonomic classifications obtained from DNA barcoding of individual specimens in each of the 45 samples.

\mysubsection{Bulk image annotation}\label{sec:bulkAnnotation}

We annotated the bulk images in two steps: first, creating segmentation masks around each visible specimen, and second, assigning taxonomic labels to each arthropod mask. This workflow confined the need for taxonomic expertise to the second step. Our approach combined three key strategies to facilitate annotation: watershed segmentation for generating initial masks \cite{schneider2022}, the Toronto Annotation Suite (TORAS) \cite{torontoannotsuite} for AI-assisted mask refinement, and the DNA barcoding-derived sample-specific taxonomies to restrict the set of suggested taxonomic labels to taxa that were confirmed to be present in each sample. The complete annotation workflow is described in detail below.

\mysubsubsection{Annotation workflow}
\label{sec:annotWorkflow}

\mysubsubsection{Step 1: Create segmentation masks}

To facilitate rapid annotation of a large number of arthropods, we used a watershed algorithm to generate initial segmentation masks \cite{schneider2022}. Watershed segmentation treats the image as a topographical surface, where pixel intensities represent heights, and finds boundaries between regions by simulating water filling up from local minima. Contiguous areas with pixel values below a threshold (200 in 8-bit grayscale) were grouped into a segmentation mask. Although computationally efficient, this simple algorithm often merged close groupings of arthropods into single masks and excluded light-coloured or translucent body parts, such as wings, and slender structures, like legs and antennae, from the masks. Therefore, we improved the masks by manually editing them using TORAS, a web-based annotation tool harnessing human-in-the-loop AI models to speed up and improve annotations for computer vision.

To permit fast annotation of tiny objects, we implemented two custom features in TORAS. First, we modified the default zoom behaviour when creating a new segmentation mask: instead of zooming out to show the full image, we maintained the current zoom level to prevent the annotator from losing track of individual insects. Second, we integrated a scale bar to help annotators better gauge object sizes within the images. Further, since the bulk samples contained between \num{36} and \num{3228} arthropods each, we used a custom script to split images into subimages for faster mask rendering in TORAS. We first split images into 4$\times$4 equally-sized subimages. To avoid splitting arthropods across subimage boundaries, we used the initial watershed masks to locate them. We calculated each mask’s centroid and assigned it to the corresponding subimage. We then adjusted subimage sizes to include the full range of each initial segmentation mask, with a 100-pixel buffer to allow for manual mask adjustments. This method of splitting resulted in some overlap, causing some arthropods to appear in multiple subimages. We visually tagged these arthropods in all but one of the subimages they appeared in to prevent redundant manually created masks.  

We uploaded the initial segmentation masks to TORAS and automatically improved them with the built-in segmentation refinement tool, which adjusts each mask to include only pixels belonging to the object. Compared to the raw watershed masks, the automatically refined masks provided a more precise fit around the arthropods. The annotators then manually edited the refined masks to ensure that each arthropod had a separate segmentation mask, capturing all its pixels without including any background. The annotators could utilize the full range of tools available in TORAS for editing the masks. For example, \texttt{Paint} and \texttt{Erase} were commonly used to manually edit the masks, while the \texttt{Box Tool}, which estimates a segmentation mask based on a bounding box provided by the annotator, was used to generate masks for arthropods missed by the watershed algorithm. Finally, the annotators assigned each segmentation mask to one of four coarse classes: arthropod (\texttt{b} for ``bug''), debris (\texttt{d}), edge (\texttt{e}, including tray edges and QR codes) or unidentifiable (\texttt{u}). For detailed annotator instructions, see \flexisuppref{sec:annotation_instructions}{ S1}.

\mysubsubsection{Step 2: Assign taxonomic labels}

In the second step of annotations, an ``expert annotator'' with experience in arthropod identification used TORAS to assign taxonomic labels to each segmentation mask previously classified as containing an arthropod (class \texttt{b}). To minimize misspellings or taxonomic disagreements between ranks, the expert annotator selected labels from the sample-specific taxonomy constructed from the individual-level DNA-based taxonomic classifications (see \flexiref{sec:inddata}{Individual specimen analysis}). However, since there was some, albeit small, proportion of specimens for which the individual DNA barcoding failed, the expert annotator was able to override the default set of choices by creating and assigning labels that did not occur in the sample-specific taxonomy. The expert annotator was asked to assign the lowest taxonomic group possible for each segmentation mask containing an arthropod.

To allow the expert annotator to convey as detailed taxonomic information as possible, we distinguished between labels of high and low confidence using two different methods. First, by assigning multiple taxonomic labels at different ranks to a single segmentation mask: the label which belonged to the highest taxonomic rank was considered high confidence, while all lower-ranking labels were treated as low confidence. That is, if the expert annotator assigned \texttt{order:} \texttt{Lepidoptera} and \texttt{family:}\texttt{Tortricidae}, Lepidoptera was considered high confidence and Tortricidae low confidence. Second, by assigning multiple labels at the same rank: their most recent common ancestor was considered a high-confidence label, while all other labels were treated as low-confidence. For example, if the expert annotator assigned both \texttt{family:} \texttt{Tortricidae} and \texttt{family:} \texttt{Geometridae}, these family labels would be considered low confidence, while their most recent common ancestor, \texttt{order:} \texttt{Lepidoptera}, would be interpreted as high confidence, regardless of whether the expert annotator explicitly assigned it. These methods allowed the expert annotator to assign uncertainty without requiring a separate step for assessing label confidence. They also avoided the difficulty of the expert quantifying their confidence.

In addition to assigning taxonomic labels, we asked the expert annotator to perform a quality check of the segmentation mask from the first annotation step. This was done with a custom feature in TORAS, which allowed annotators to view the segmentation masks in the second annotation step. During the quality check, the expert annotator primarily adjusted the mask if visual characteristics important for the classification of the arthropod, such as wings or antennae, were not included in the mask, or if multiple arthropods were grouped in one mask. The expert annotator could also change the initial assignment to one of the four coarse classes, if, for example, debris or an unidentifiable object had mistakenly been labelled as an arthropod. For detailed annotation instructions, see \flexisuppref{sec:expert_instructions}{ S1.4}.

\begin{figure*}[tb!]
	\centering    
	\includegraphics[width=0.9\textwidth]{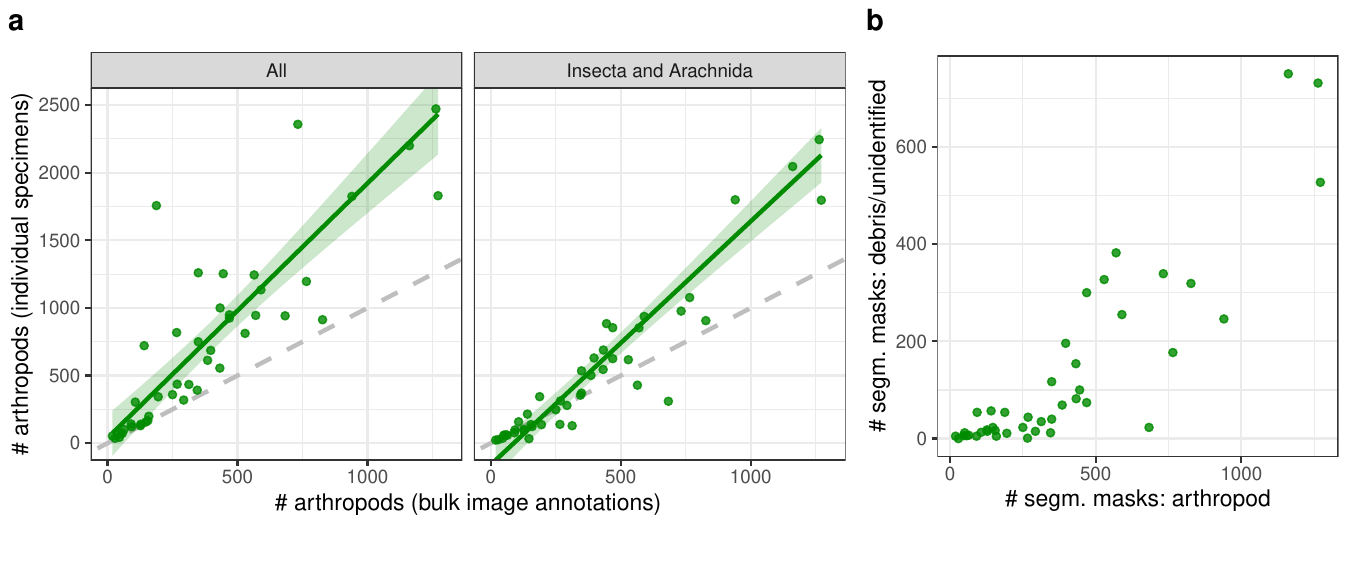}
 \caption{(\textbf{a}) For each sample, a comparison was made between the number of arthropods annotated in the bulk images and the number of individual specimens isolated from the corresponding samples, here shown for all taxa (left) and restricted to Insecta and Arachnida (right). The green line represents a linear regression fit with a 95\% confidence interval between the two quantities, and the dashed grey line indicates a 1:1 relationship. (\textbf{b}) For each sample, a comparison was made between the number of segmentation masks tagged as arthropods and the number tagged as debris or unidentifiable across all bulk images.}
 \label{fig:annotstatus}
\end{figure*}

\mysubsubsection{Annotation completeness} \label{sec:annot-completeness}

To evaluate how accurately the bulk image annotations reflect the true number of arthropods in the samples, we compared the number of segmentation masks annotated as arthropods with the actual number of specimens isolated from each sample (Figure \ref{fig:annotstatus}a). We found that in samples containing more than approximately \num{250} arthropods, the number of arthropods based on the bulk image annotations was substantially lower than the true count. Some of these discrepancies occurred in samples with a high abundance of springtails (Collembola), which are often small, pale, and difficult to separate from debris in the bulk samples. Restricting the comparison to individual specimens classified as Insecta or Arachnida (both of which are typically larger and darker than springtails) reduced the difference between annotated counts and true specimen counts. Overall, the absolute discrepancy in counts tended to increase with larger samples, suggesting two possibilities. First, it is inherently challenging for human annotators to detect all insects in images where the total number of individuals is very high. These samples also often contained substantial debris, which can obscure smaller insects (Figure \ref{fig:annotstatus}b). Additionally, because each insect occupies only a small proportion of the image, especially tiny insects may appear visually indistinct or blurry, making them difficult to annotate correctly. Second, annotator fatigue may set in for these large samples, leading to fewer corrections for arthropods missing a segmentation mask once the total count is already high. 

We were able to annotate the majority of specimens in the bulk images at rank suborder or above with high confidence (Table \ref{tab:annots_rank}). Including low-confidence annotations increased the number of specimens annotated at lower ranks, with almost half of the specimens annotated at superfamily level, and more than a third at the family level (Table \ref{tab:annots_rank}).

\begin{figure}
    \centering
    \includegraphics[width=1\linewidth]{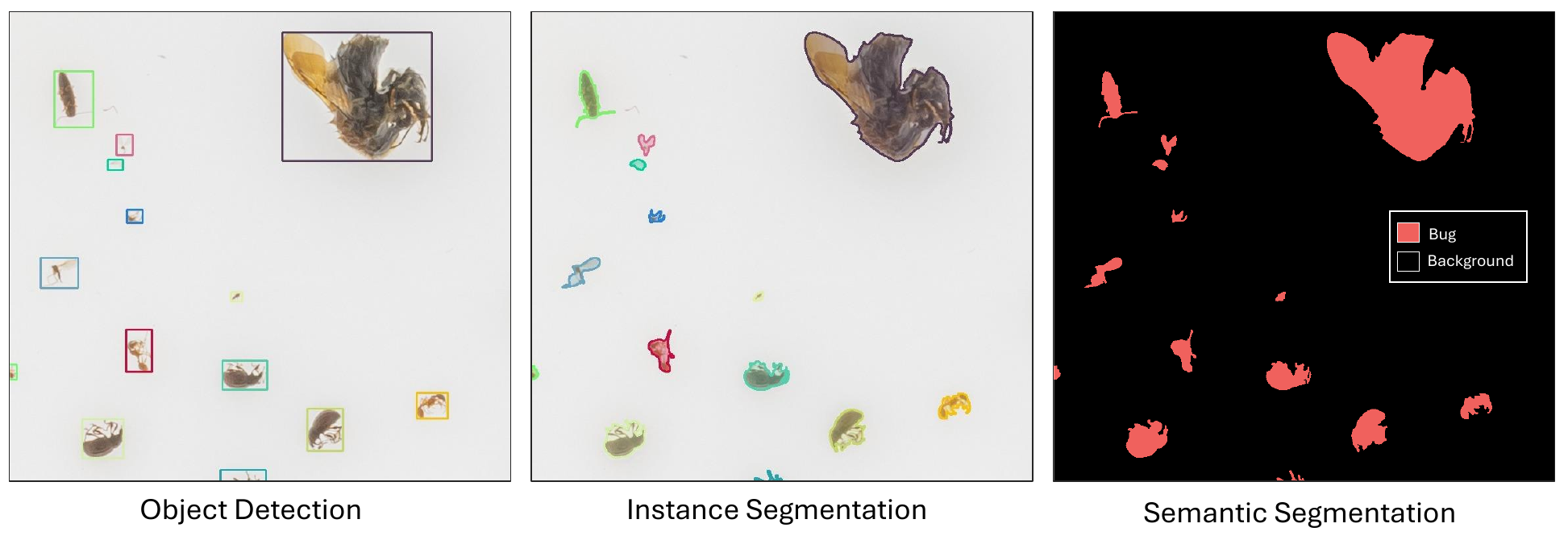}
    \caption{Differences between annotations for object detection (left), instance segmentation (middle), and semantic segmentation (right).}
    \label{fig:obj-det-vs-seg}
\end{figure}

\begin{table}
\centering
\caption{Number of unique taxa and specimens annotated at each taxonomic rank. Values are showed separately for high-confidence annotations (HC) and low-confidence annotations (LC).}
\label{tab:annots_rank}
\begin{tabular}{lccccc}
  \toprule
Rank & \# Taxa & Labelled\textsubscript{HC} & Labelled\textsubscript{LC} & Labelled\textsubscript{HC} (\%) & Labelled\textsubscript{LC} (\%) \\ 
  \midrule
Phylum & 1 (1) & \num{17892} & \num{17892} & 100.0\%  & 100.0\% \\ 
  Class & 4 (4) & \num{17570} & \num{17834} & 98.2\% &  99.7\% \\ 
  Order & 23 (25) & \num{15042} & \num{16463} & 84.1\% &   92.0\% \\ 
  Suborder & 8 (8) & \num{11218} & \num{12745} & 62.7\% &  71.2\% \\ 
  Infraorder & 5 (6) & \num{7792} &  \num{10778} &   43.5\% & 60.2\% \\ 
  Superfamily & 34 (47) & \num{6261} & \num{8730} &   35.0\% & 48.8\% \\ 
  Family & 92 (129) & \num{4546} & \num{6358} & 25.4\% &  35.5\% \\ 
  Subfamily & 27 (36) & \num{993} & \num{1174} & 5.5\%  &  6.6\% \\ 
  Genus & 35 (55) & \num{584} & \num{694} & 3.3\%  &  3.9\% \\ 
  Species & 17 (23) & \num{63} & \num{74} & 0.4\%  &  0.4\% \\ 
   \bottomrule
\end{tabular}
\end{table}

\mysubsection{Machine learning dataset}\label{sec:ml_dataset}

MassID45 serves as a benchmark dataset for instance segmentation of tiny, densely packed objects. Instance segmentation differs from other tasks like object detection, which uses bounding boxes, and semantic segmentation, which labels pixels but does not distinguish between individual objects. In contrast, instance segmentation assigns a unique pixel-wise mask to each individual object (Figure \ref{fig:obj-det-vs-seg}). Using the fully annotated set of 49 bulk images, we limited the task to instance segmentation of arthropod specimens (class \texttt{b}), i.e., excluding objects tagged as debris or unknown. This allowed us to approach this problem as a single-class, small-instance segmentation task.

\begin{figure}[t!]
    \centering
    \includegraphics[width=1\linewidth]{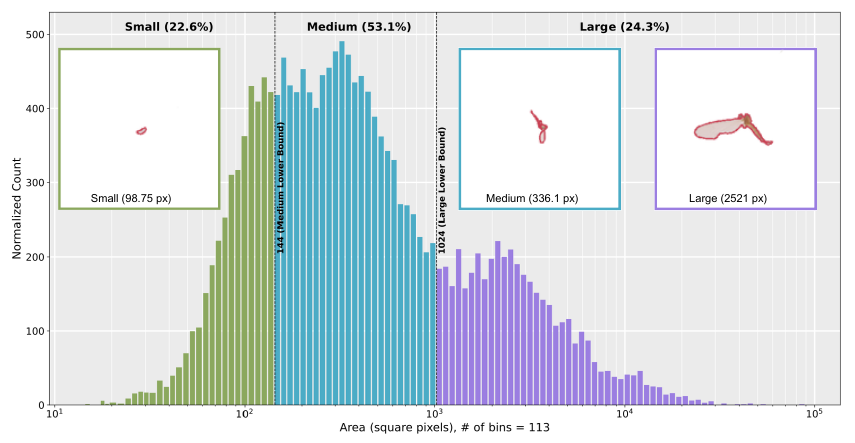}
    \caption{Distribution of insect mask areas for ``small'' ($<\!144$ pixels), ``medium'' ($\geq{\!144}$ but $<\!1024$ pixels), and ``large'' ($\geq{\!1024}$ pixels) insects. Counts are adjusted such that the area of a bar is proportional to the count in that bin. The three images show the median masks for small, medium, and large insects, all at the same magnification.}
    \label{fig:area_histograms}
\end{figure}

    Annotations are exported from TORAS in the same format as the Microsoft Common Objects in Context (MS-COCO) dataset \citep{MS-COCO}, a benchmark dataset for pretraining object detection and instance segmentation models. We based the evaluation scheme for the instance segmentation models on MS-COCO conventions, including metrics that rely on the areas of the instance masks to be detected. However, the MS-COCO evaluation scheme was designed for objects that are significantly larger than those in the MassID45 dataset, with \num{76.5}\% of the arthropod masks being categorized as ``small''. Therefore, to increase the granularity in our performance evaluations, we instead used the area thresholds from iSAID \cite{zamir_isaid_2019}, a dataset of remote sensing images intended for small object detection and instance segmentation. Thus, we defined object sizes as follows: ``small'' for areas $<\!144$ pixels; ``medium'' for areas $\geq{\!144}$ but $<\!1024$ pixels; and ``large'' for areas $\geq{\!1024}$ pixels. In total, the 49 fully annotated bulk images contain segmentation masks for \num{17937} arthropods. Mask areas range between \num{15.1} and \num{83182.4} pixels, with a mean and median of \num{1152.2} and \num{343.4} pixels, respectively (Figure \ref{fig:area_histograms}).

\mysubsection{Preprocessing bulk images}\label{sec:preprocessing}

Before training deep neural networks to perform instance segmentation, we preprocessed the image and segmentation mask data. We first merged the annotations from the subimages back together to match the original bulk images (\flexiref{sec:bulkAnnotation}{Bulk image annotation}). This allowed us to break up the images into tiles as needed for model training, detailed below. We used the Shapely library \cite{Gillies_Shapely_2025} in Python to merge segmentation masks with multiple polygons and to correct invalid (i.e., self-intersecting) polygon masks. The insect masks were processed as concave hulls, filling in holes (e.g., areas between legs) in the edited segmentation masks to create single polygons. For \num{63} of the \num{17937} insect masks (0.351\%), these preprocessing steps resulted in several unconnected polygons that could not be merged via a unary union. In such cases, we took the polygon with the largest area as the final mask. This ensured that deep learning models only needed to predict one polygon per annotation, simplifying the segmentation task. After cleaning the segmentation masks, we manually cropped the bulk images to only contain the areas in which insects were present. This resulted in finalized cropped bulk images of different dimensions.

\mysection{Data records} 

The MassID45 dataset is organized into two resolution levels (Table \ref{tab:data_overview}): bulk samples containing bulk images,  metabarcoding data, and taxonomic image annotations and individual specimens containing individual images and DNA barcoding data. Sample metadata, bulk sample images, bulk image annotations, and models described here are all available from Zenodo \cite{MassID45-Zenodo}. Sample metadata is provided in a CSV file with one row per sample, uniquely identified by a six-character alphanumeric code. The same sample code is used as the file name of the corresponding bulk image, followed by suffix \texttt{\_\{image\}}, where image is 1 or 2, in cases where there is more than one image per sample. We provide each image raw in CR3 format and edited in JPEG format. Bulk image annotations are available from step 1 and 2 in both COCO and TORAS format. The trained models are provided as PyTorch checkpoints and can be used for model inference with the code provided at \url{https://github.com/uoguelph-mlrg/MassID45}. 

Raw sequencing reads for bulk samples are available from ENA \cite{MassID45-ENA} under project accession number PRJEB86111. The sequences for each sequencing replicate are represented by two gzipped FASTQ files, containing the R1 and R2 paired-end reads.  Thus, for each physical sample there are a total of six files, with names of the form \texttt{\{sample\}\_Rep\{i\}.R\{read\}.fastq.gz}, where \texttt{\{sample\}} is the six-character alphanumeric code uniquely identifying the sample, \texttt{\{i\}} is an integer between 1 and 3 indicating the replicate number, and \texttt{\{read\}} is 1 or 2.
Accession numbers for individual samples and read files, along with a script to download all relevant files, are provided in \texttt{MassID45\_ENA\_accnos.tsv}, and \texttt{download\_MassID45\_ENA.sh}, respectively, both available at the above GitHub repository.
Individual arthropod images and DNA barcode sequences are available as project ID \texttt{DS-LPEPA22} on BOLD \cite{MassID45-BOLD}. On BOLD, the \texttt{field ID} variable corresponds to the sample code used in the sample metadata and bulk image names, while the \texttt{sample ID} is an identifier unique to each individual specimen.

\begin{table}
\centering
\caption{Overview of data types included in the MassID45 dataset.}
\label{tab:data_overview}
\small
\begin{tabular}{llll}
\toprule
\textbf{Resolution} & \textbf{Data type} & \textbf{Quantity} & \textbf{Description} \\
\midrule
{\makecell[l]{Bulk samples \\ $N=45$}}
    & Bulk images & \makecell[l]{49 images\\(of 45 samples)} & \makecell[l]{Images depicting unsorted insect samples, \\ with 1-2 images per sample (41 samples 1:1, 4 samples 1:2).} \\
    & Metabarcoding data & 45 samples & \makecell[l]{COI sequences from metabarcoding of unsorted insect \\ samples. Each sample has three technical replicates.} \\
    & Taxonomic image annotations & \num{17940} annotations & \makecell[l]{Segmentation masks and expert taxonomic assignments \\ for individual arthropods in the bulk images.} \\
\midrule
{\makecell[l]{Individual specimens \\ $N=$ \num{35586}}}
    & Individual images & \num{35586} images & \makecell[l]{Images of each arthropod specimen from \\ the 45 bulk samples.}\\
    & Barcoding data & \num{35586} sequences & \makecell[l]{COI sequences from DNA barcoding of \\ individual insect specimens.} \\
\bottomrule
\end{tabular}
\end{table}

\mysection{Technical validation}\label{sec:tech-valid}

In this work, we benchmark instance segmentation performance on MassID45 using two paradigms: zero-shot learning and supervised learning. This analysis allows us to evaluate how valuable the expert annotations are for detecting small arthropods, compared to ``out-of-the-box'' generalist models. 

Under the zero-shot paradigm, we employ models that have not seen any training examples from the MassID45 data. Zero-shot models rely exclusively on their pre-training data -- which includes large, diverse computer vision datasets -- to generalize to images from unseen domains \citep{wang_cut_2023, liu2023grounding, ren2024grounded, xiao2023florence, ravi2024sam2segmentimages, gemini_flash_2}. With supervised learning, we instead train instance segmentation models \citet{he_mask_2018, li_mask_2022, cheng_masked-attention_2022} using annotated examples from the MassID45 dataset. By comparing the performance of zero-shot and supervised approaches, we can assess whether the expert annotations are valuable enough to justify the annotation effort, or whether existing generalist models achieve adequate detection performance on the MassID45 data. We describe implementation details for our training and inference pipelines below. 


\mysubsection{Experimental setup}

\mysubsubsection{Dividing bulk images into tiles}\label{sec:sahi}
Due to GPU memory constraints and the high resolution of the images, we could not present entire bulk images to deep learning models during training or inference. As a solution, we split the bulk images into tiles, similar to previous work \citep{unel_power_2019, ding_object_2022}. Compared to down-sampling the images, which can also be used to produce a resolution which fits within GPU memory, tiling preserves the pixel density of the original images. Using tiling thus avoids a loss of visual details, which is particularly important for the small insects in the MassID45 dataset. We used a sliding window to crop tiles out from the bulk images, and each tile was then treated as a separate image during model training and inference. We determined the optimal tile size for training and inference to be $\num{512} \times \num{512}$ pixels (see \flexisuppref{sec:upsampling-factor}{ S2.1}). During tiling, some insects may get cut between tiles. To mitigate this, we used an overlap of \num{60}\% between tiles, similar to previous work on small object detection \cite{ding_object_2022}, thus ensuring cut insects along the boundary of one tile were shown intact in adjacent tiles.

Tiling introduces a challenge during inference: when the same insect appears in multiple overlapping tiles, treating each tile as an independent image would lead to duplicate detections and inaccurate performance estimates. To address this, we implemented slicing-aided hyper-inference (SAHI), a method designed to merge predictions across overlapping tiles and accurately reconstruct detections in the full bulk image \cite{akyon_slicing_2022}. The SAHI algorithm has previously been used for small object detection problems in remote sensing \cite{nguyen_yosca_2024, lin_stpm_sahi_2022, gia_enhancing_2024} and pest monitoring \cite{fotouhi_persistent_2024}. We used SAHI to postprocess the tiled predictions by applying non-maximum merging (NMM) \citep{kondrackis_nmm_2024}. NMM relies on the Intersection over Union (IoU; see \flexiref{sec:metrics}{Evaluation metrics} below), a measure of how much two masks overlap, to identify and merge predictions which are likely duplicates. After sorting predicted masks across all tiles by their confidence scores, NMM identifies and groups detections that overlap by more than a predetermined IoU threshold ($\mathit{IoU}_\mathit{NMM}$). Within each group of overlapping masks, NMM iteratively merges pairs of predictions, producing a new mask that spans their combined area and a new confidence score weighted by the original masks' confidence scores and areas. This pairwise merging continues until one mask remains for each group of overlapping detections. Lastly, we mapped the final set of merged predictions from the tiles back onto the original bulk image, allowing us to evaluate them directly against the ground truth bulk images. When merging the predictions across tiles, we used an $\mathit{IoU}_\mathit{NMM}$ of 50\%, meaning that overlapping predictions were considered duplicates and iteratively merged if their intersection over union was at least 50\%.

\mysubsubsection{Data partitioning}

We randomly partitioned the bulk images into training (40 images, $81.6\%$), validation (3 images, $6.1\%$), and testing (6 images, $12.2\%$) sets. After dividing the bulk images into $512 \times 512$ tiles, this resulted in \num{17062} training tiles, \num{1244} validation tiles, and \num{1586} testing tiles. To prevent data leakage, all tiles from a given bulk image were assigned to the same dataset split. Including insects that were duplicated and/or partially cut between tiles, the tiled training set contained \num{110520} insects, the tiled validation set \num{5867}, and the tiled test set \num{6241}. The validation and test sets contained data not seen during training and served to evaluate how well the models generalized to new, real-world data. The validation set was used to guide intermediate modelling decisions, such as selecting between models or preprocessing techniques, while the test set was used to measure the performance of the final model after all model development and experimentation was complete.

\mysubsubsection{Data augmentations}\label{sec:data-augs}

To artificially increase the number of training samples and improve generalization, we applied data augmentations to the tiled images from the training partition, drawing on prior work focused on small object detection in remote sensing and underwater imagery \cite{ding_object_2022, galloway_predicting_2022}. It is important to note that our tiling process also acted as a form of data augmentation, as the arthropods could be present in multiple adjacent tiles. We employed both geometric and colour-based augmentations (Table \ref{tab:data_augs}), which introduced variations to the bulk images while ensuring the insects could still be identified. For example, random rotations and horizontal flips mimicked the possible orientations that arthropods can assume when placed in the sorting trays. Random adjustments to brightness, contrast, and saturation were intended to make the models more robust to small differences in lighting across bulk images, as well as natural colouration differences among arthropods (e.g., in different life stages). We applied these augmentations to the tiled bulk images, then resized each augmented tile to a fixed input size of $1024 \times 1024$ using bilinear interpolation before presenting them to the model during training. 

\begin{table}
\centering
\caption{Geometric and colour-based data augmentations used for the training data, where $p$ denotes the probability of applying each transformation.}
\label{tab:data_augs}
\begin{tabular}{lll}
\toprule
\textbf{Category} & \textbf{Augmentation} & \textbf{Parameters} \\
\midrule
{Geometric}
    & Random horizontal flip & $p=0.5$ \\
    & Random rotation & \{0°, 90°, 180°, 270°\}, $p=0.25$ each \\
\midrule
{Colour}
    & Random brightness & Uniform in range $[-15\%, +15\%]$ \\
    & Random contrast & Uniform in range $[-10\%, +10\%]$ \\
    & Random saturation & Uniform in range $[-15\%, +15\%]$ \\
\bottomrule
\end{tabular}
\end{table}

\mysubsubsection{Evaluation metrics}\label{sec:metrics}
Using the predictions merged with SAHI, we calculated evaluation metrics following the MS-COCO evaluation scheme \cite{MS-COCO}, which relies on IoU, precision, and recall. For a given instance mask prediction, IoU quantifies the overlap between the predicted instance masks and ground truth annotations:
\begin{equation*}
\text{IoU} = \frac{\mathit{TP}_p}{\mathit{TP}_p + \mathit{FP}_p + \mathit{FN}_p},
\end{equation*}
where $\mathit{TP}_p$ represents the number of predicted pixels that matched the ground truth (true positives), $\mathit{FN}_p$ denotes the number of ground truth pixels missed by the prediction (false negatives), and $\mathit{FP}_p$ represents the number of background pixels incorrectly labelled as part of the instance (false positives). When calculating the evaluation metrics, we used a confidence threshold ($\mathit{conf}_\mathit{eval}$) to filter out uncertain predictions and an IoU threshold ($\mathit{IoU}_\mathit{eval}$) to define how strictly the predicted masks must overlap with the ground truth annotations to be considered correct. That is, we categorized each predicted instance as a true positive ($\mathit{TP}_i$), false positive ($\mathit{FP}_i$), or false negative ($\mathit{FN}_i$) based on whether its IoU with the ground truth masks exceeded $\mathit{IoU}_\mathit{eval}$. Based on the instance-level categorizations, we then calculated precision and recall.  

Precision was calculated as 
\[\text{Precision} = \frac{\mathit{TP}_i}{\mathit{TP}_i+\mathit{FP}_i}.\] It quantifies how many of the insects detected by the model were actually correct. Conversely, recall was calculated as \[\text{Recall} = \frac{\mathit{TP}_i}{\mathit{TP}_i+\mathit{FN}_i}.\] It reflects how many of the actual insect specimens were detected by the model. We calculated precision-recall curves by keeping $\mathit{IoU}_\mathit{eval}$ fixed and varying $\mathit{conf}_\mathit{eval}$. Following the MS-COCO evaluation scheme \citep{MS-COCO}, we then calculated the average precision (AP), defined as the area under the precision-recall curve, for several $\mathit{IoU}_\mathit{eval}$ thresholds. Here, we report the following aggregate metrics:  
\begin{itemize}
    \item AP\textsubscript{50:5:95}: mean of the AP  values calculated across $\mathit{IoU}_\mathit{eval}$ thresholds ranging from \num{50}\% to \num{95}\% in \num{5}\% increments.
    \item AP\textsubscript{50}: AP at a fixed $\mathit{IoU}_\mathit{eval}$ of 50\%.
    \item AP\textsubscript{75}: AP at a fixed $\mathit{IoU}_\mathit{eval}$ of 75\%.
\end{itemize}
We also measured AP\textsubscript{50:5:95} for the ``small'', ``medium'', and ``large'' object categories (\flexiref{sec:ml_dataset}{Machine learning dataset}), denoting them as AP\textsubscript{S}, AP\textsubscript{M}, and AP\textsubscript{L}, respectively. We report the final evaluation metrics for each supervised baseline on the test set of six bulk images. 

\mysubsection{Benchmarking instance segmentation models}
\begin{table}
\centering
\caption{Instance segmentation results on the MassID45 test set for the zero-shot and supervised baselines. For each mask AP metric, the top result per paradigm is \textbf{bolded}.}
\vspace{2mm}
\label{tab:detector_comparison}
\begin{tabular}{llcccccc}
\toprule
Detector paradigm & Detector & AP\textsubscript{50:5:95}& AP\textsubscript{50} & AP\textsubscript{75} & AP\textsubscript{S}& AP\textsubscript{M}& AP\textsubscript{L}\\
\midrule
{Zero-shot}
& CutLER & 22.7& 40.0 & 22.1 & 0.80 & 18.1 & 59.0 \\
& Grounding DINO + SAM 2.1 & \textbf{27.1}& 47.6 & \textbf{27.0} & 1.30 & \textbf{22.6} & \textbf{66.3} \\
& Florence-2 + SAM 2.1 & 16.5& 28.8 & 16.7 & 3.00 & 12.1 & 41.7 \\
& Gemini 2.0 Flash + SAM 2.1 & 26.2& \textbf{50.0}& 23.8 & \textbf{3.30} & 18.3 & 64.2 \\
\midrule
{Supervised}
&Mask R-CNN & 42.5& \textbf{83.1}& 36.6& 20.0& 41.6& 70.4\\
&Mask2Former & 41.4 & 78.7 & 37.4 & 20.5 & 40.0 & 71.1 \\
&Mask DINO & \textbf{43.5}& 80.9 & \textbf{40.1}& \textbf{21.1}& \textbf{43.5}& \textbf{73.1}\\
\bottomrule
\end{tabular}
\end{table}

\begin{figure}
    \centering
    \includegraphics[width=\linewidth]{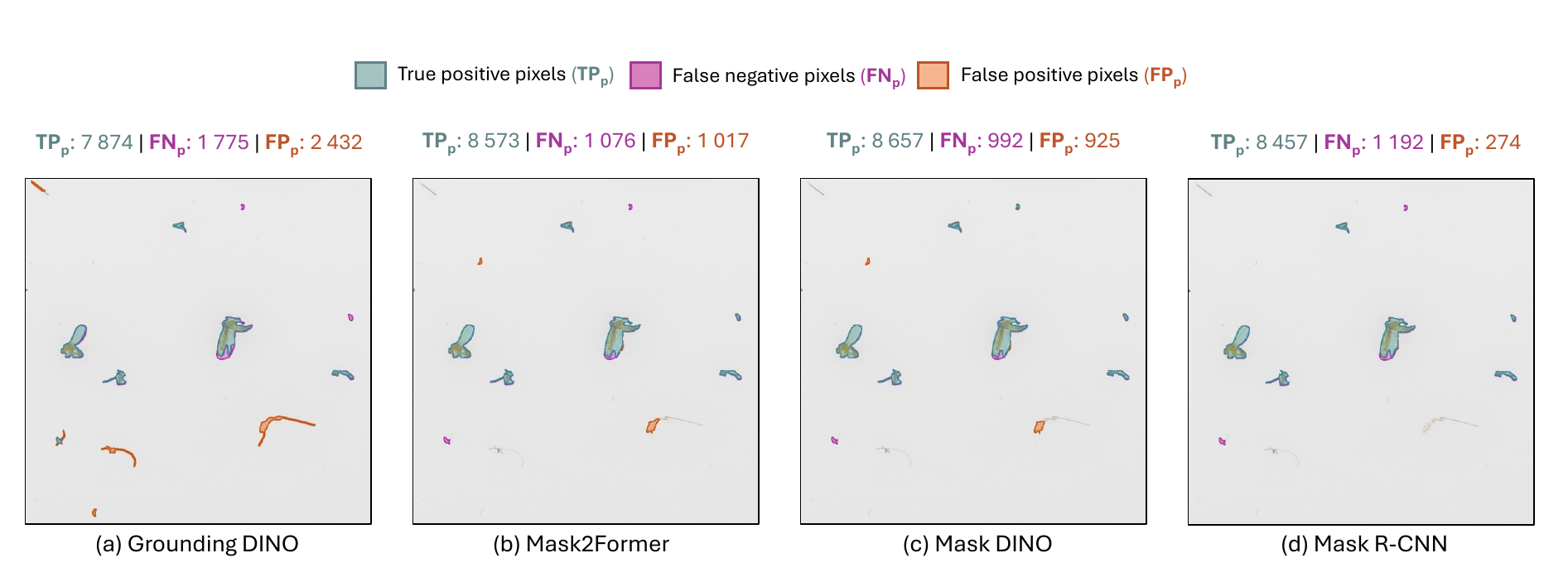}
    \caption{Visual instance segmentation results for one example patch from the MassID45 test set. Predicted masks are compared for (a) the top-performing zero-shot model, Grounding DINO; and (b)~--~(d) the 3 supervised baselines: Mask2Former, Mask DINO, and Mask R-CNN. Above each panel, we show the areas occupied by TPs, FPs, and FNs in pixels. Best viewed on a colour display with zoom.}
    \label{fig:example-detections}
\end{figure}
\mysubsubsection{Implementing zero-shot detectors}

Zero-shot approaches can localize objects from a new domain without any prior fine-tuning on that domain, relying exclusively on pretraining from large, diverse datasets --- including multi-modal data. We benchmark their generalization capabilities by applying them to a challenging new domain: small arthropods from the MassID45 data. For consistency, we applied the same SAHI approach using  $512 \times 512$ pixel tiles with \num{60}\% overlap (see \flexiref{sec:sahi}{Merging Predictions across Tiles}). 

We selected methods representing different forms of zero-shot detection, including unsupervised instance segmentation (CutLER) \cite{wang_cut_2023}, open-vocabulary or open-set models that use text prompts (Grounding DINO \cite{liu2023grounding, ren2024grounded}, and Florence-2 \cite{xiao2023florence}), as well as large, state-of-the-art, multi-modal models (Gemini 2.0 Flash) \cite{gemini_flash_2}. To perform instance segmentation, we paired the latter three methods with Meta’s Segment Anything Model 2 (SAM~2.1) \cite{ravi2024sam2segmentimages}, a foundation model for image segmentation. The bounding boxes from Grounding DINO, Florence-2, and Gemini 2.0 Flash were used as prompts for SAM~2.1, producing instance masks that were used in our evaluation scheme for instance segmentation. For implementation details --- including the model checkpoints and text prompts used for the zero-shot models --- see \flexisuppref{sec:zero-shot}{ S2.2}.

\mysubsubsection{Implementing supervised detectors}\label{sec:supervised-experiments}

For the supervised models, we selected three general architectures originally developed for standard computer vision datasets like MS-COCO \cite{MS-COCO}, which includes millions of images of everyday objects. Here, we seek to determine whether they can be adapted to small, detailed organisms like arthropods in MassID45 when guided by expert annotations. These models include a popular baseline for instance segmentation, Mask R-CNN \cite{he_mask_2018, wu2019detectron2}, and two more recent methods, Mask2Former \cite{cheng_masked-attention_2022} and Mask DINO \cite{li_mask_2022}. The latter two use transformer-based architectures, an approach that has driven recent advances in computer vision \citep{shehzadi2023objectdetectiontransformersreview, rekavandi2023transformerssmallobjectdetection, li2024transformerbasedvisualsegmentationsurvey}, to achieve state-of-the-art results on the MS-COCO benchmark.

All supervised models were initialized with weights from a ResNet-50 backbone pretrained on the MS-COCO dataset \cite{MS-COCO}, allowing us to leverage features from a large benchmark dataset. Although MS-COCO contains no arthropods, it includes nearly 1.5 million labelled object instances, which provides models with general-purpose visual features such as edges, shapes, textures, and colour patterns. These foundational features can then be transferred across domains, like fine-grained biological imagery. This strategy, known as transfer learning \cite{bozinovski_reminder_2020}, offers a practical alternative to training models from scratch (i.e., random weights). Using MS-COCO pretrained checkpoints for instance segmentation ensures that all three models start with the same baseline of visual understanding, allowing us to more fairly compare how each architecture adapts to the specialized task of segmenting small arthropods in the MassID45 dataset. Training instead with randomly initialized weights would require the models to learn all visual features from a comparatively small dataset, increasing the risk of poor generalization.

Using the Detectron2 library \citep{wu2019detectron2}, we fine-tuned each model for \num{15000} iterations with a batch size of \num{8} (\num{2} images per GPU with \num{4} GPUs), using the AdamW \citep{AdamW} optimizer with a peak learning rate of $5\times10^{-5}$ and weight decay of \num{0.05}. In all training runs, we used a one-cycle cosine annealed learning rate schedule \citep{onecycle} with a warm-up period of \num{4500} iterations. Training was performed using four NVIDIA RTX6000 GPUs. For inference, we applied the SAHI (\flexiref{sec:sahi}{Merging Predictions across Tiles}) approach, dividing the bulk images from the test partition into $512 \times 512$ pixel tiles with \num{60}\% overlap. We then used an $\mathit{IoU}_\mathit{NMM}$ of \num{50}\% to map the predictions from the tiles back to the original bulk image dimensions. 

\mysubsubsection{Performance evaluation} 

Without fine-tuning on the MassID45 data, the zero-shot models performed significantly worse than their supervised counterparts (Table \ref{tab:detector_comparison}). Grounding DINO, the top-performing zero-shot method, only achieved a mask AP\textsubscript{50:5:95} of 27.1\%, which is far below the 43.5\% achieved by the Mask DINO, the best model across almost all AP metrics. 

It is important to note that the reported AP evaluation metrics describe performance across several IoU and confidence thresholds. When these instance segmentation models are deployed in a real-world setting, we must select a fixed operating point for the confidence threshold. For each detector, we selected distinct confidence thresholds that maximized that model's F1-score --- the harmonic mean between precision and recall --- on the validation set (see \flexisuppref{sec:confidence-thresholds}{ S2.3}). We then used these confidence thresholds to filter our predictions on the test. Using these fixed confidence thresholds, we visualized predictions from each model on an exemplar patch from the test set (Figure \ref{fig:example-detections}). 

Qualitatively, Grounding DINO could successfully localize and segment larger arthropods, but missed most small insects. It also misidentified QR codes as insects (Figure \ref{fig:example-detections}a). In contrast, the supervised models produced instance masks that align well with the ground truth. However, we observed that separating debris from small insects was a difficult task, as the supervised models had a tendency to confuse small, loose debris with insects and vice-versa (Figure \ref{fig:example-detections}b-d). For this exemplar patch, we also reported the number of TP, FP, and FN pixels to illustrate the differences between each model's predictions. The zero-shot Grounding DINO model predicted significantly more FPs and FNs than the supervised models. Conversely, the three supervised models predicted similar numbers of FPs and FNs, with Mask DINO predicting the fewest FNs, and Mask R-CNN detecting the fewest FPs. Similar trends can be seen when aggregating the TP, FP, and FN pixels across the six bulk images in the MassID45 test set (see Table \ref{tab:segmentation-areas} in \flexisuppref{sec:confidence-thresholds}{ S2.3}). 

The relatively poor performance of the zero-shot baselines suggests fine-tuning is still needed for specialized tasks like detecting arthropods from the MassID45 dataset. More importantly, this finding underscores the importance of expert annotations for bulk image analyses. The complexities of the detection task are caused by the small size of the arthropods, as well as their high similarity to surrounding debris. While not explored in this work, fine-tuning these zero-shot methods on the MassID45 dataset may prove beneficial. It is important to note, however, that the supervised models explored in this work are optimized for bulk images obtained using our experimental setup, and may need to be further fine-tuned on bulk images taken from different experimental conditions. Thus, this analysis frames MassID45 as a challenging benchmark dataset for custom supervised models, vision foundation models, and other zero-shot detectors, as it assesses their ability to recognize tiny, ambiguous objects rather than larger common objects that are typically considered in the literature.

\mysection{Usage Notes}

Our annotation workflow consisted of two separate steps, where only a subset of the annotations from step 1 (those categorized as arthropods, \texttt{b}) were annotated in step 2. In step 2, the main task of the annotator was to provide a taxonomic label for each specimen. However, if the second annotator disagreed with the first categorization, they could change it to one of the three other categories (\texttt{d}, \texttt{e}, or \texttt{u}). For the full set of annotations with both broad categories and taxonomic annotations, the output from step 1 and 2 must therefore be merged. For the taxonomic annotations, we used multiple labels as a way to express annotator uncertainty. If a single label is required, we therefore recommend careful selection of which taxon name to use.

While efforts were made to ensure the bulk images were fully annotated, some insects that were at the boundaries of the $\num{4} \times \num{4}$ annotator patches may have been missed. As mentioned above in \flexiref{sec:annot-completeness}{Annotation Completeness}, insects may appear blurry in the images. This limitation to image quality can be addressed by techniques like super-resolution, which reconstructs plausible high-quality details from low-resolution images. We leave this for future work.

While effective on the bulk images from the MassID45 data, the fine-tuned instance segmentation models provided in this work may not generalize to bulk images taken under different imaging protocols. Such a distributional shift would necessitate transfer learning on the user's own set of bulk images. Nevertheless, pre-trained weights from our instance segmentation models may prove beneficial for other detection tasks involving small objects. We encourage further experimentation on the MassID45 dataset, particularly with existing instance segmentation models and vision foundation models.   

\mysection{Code Availability}
The code for generating the machine learning dataset and replicating our experiments is available at \url{https://github.com/uoguelph-mlrg/MassID45}.

\mysection{Author Contributions}
J.O: Conceptualization, Methodology, Supervision, Visualization, Writing – original draft; J.Q: Formal analysis, Methodology, Writing – original draft; H.A.: Investigation, Writing – review \& editing; G.B.: Investigation, Writing – review \& editing; N.C.: Conceptualization, Methodology, Supervision, Writing – review \& editing; J.D.: Data curation, Investigation, Methodology, Project administration, Resources, Writing – review \& editing; S.D.: Data curation, Investigation, Methodology, Project administration, Resources, Writing – review \& editing; A.F.: Investigation, Writing – review \& editing; B.F.: Conceptualization, Data curation, Methodology, Supervision, Writing – review \& editing; B.H.: Project administration, Data curation, Writing – review \& editing; N.I.: Investigation, Writing – review \& editing; A.K.: Software, Writing – review \& editing; O.K.: Investigation, Writing – review \& editing; D.K.: Project administration, Data curation, Writing – review \& editing; E.K.: Investigation, Writing – review \& editing; J.M.: Data curation, Investigation, Methodology, Project administration, Resources, Writing – review \& editing; T.M.: Data curation, Methodology, Writing – review \& editing; E.N.: Investigation, Writing – review \& editing; H.R.: Investigation, Writing – review \& editing; T.R.: Conceptualization, Funding acquisition, Methodology, Supervision, Writing – review \& editing; P.S.: Investigation, Writing – review \& editing; J.S.: Data curation, Investigation, Methodology, Project administration, Resources, Writing – review \& editing; M.S.: Investigation, Writing – review \& editing; A.T.: Investigation, Writing – review \& editing; E.V.Z.: Data curation, Investigation, Methodology, Project administration, Resources, Writing – review \& editing; I.Z.: Visualization, Writing – review \& editing; A.G.: Methodology, Supervision, Validation, Writing – review \& editing; S.C.L.: Conceptualization, Methodology, Supervision, Writing – review \& editing; G.W.T.: Conceptualization, Funding acquisition, Methodology, Project administration, Supervision, Writing – review \& editing.

\mysection{Competing Interests}
The authors declare no competing interests. 

\mysection{Acknowledgements} 
The generation of samples and the contributions of JO, GB, AF, BH, DK, TM, EN, HR, and PS were funded by the European Research Council (ERC) under the European Union’s Horizon 2020 research and innovation programme (grant agreement No 856506; ERC-synergy project LIFEPLAN). Barcoding, metabarcoding and imaging of the samples were funded by the Swedish Environmental Protection Agency (agreement 225-20-002 with TR). JQ, IZ, AG, SCL were supported by the \href{http://abcresearchcenter.org/}{AI and Biodiversity Change (ABC) Global Center}, which is funded by the US National Science Foundation under \href{https://www.nsf.gov/awardsearch/showAward?AWD_ID=2330423&HistoricalAwards=false}{Award No. 2330423} and Natural Sciences and Engineering Research Council of Canada under \href{https://www.nserc-crsng.gc.ca/ase-oro/Details-Detailles_eng.asp?id=782440}{Award No. 585136}. We also acknowledge the support of the Government of Canada's New Frontiers in Research Fund \href{}{Award No. NFRFT-2020-00073}. Resources used in preparing this research were provided, in part, by the Province of Ontario, the Government of Canada through CIFAR, and  \href{https://www.vectorinstitute.ai/partnerships/}{companies sponsoring} the Vector Institute.


\bibliography{main}

{%
  \renewcommand{\thefigure}{S\arabic{supsection}.\arabic{suppfigure}}%
  \renewcommand{\thetable}{S\arabic{supsection}.\arabic{supptable}}%

  \clearpage

\setcounter{page}{1}

\section*{SUPPLEMENTARY INFORMATION}\label{sec:supplementary}
\supsection{Annotation instructions}\label{sec:annotation_instructions}

\newcommand{\opus}[1]{%
  \begingroup
    \spaceskip=\fontdimen2\font plus \fontdimen3\font minus \fontdimen4\font
    \xspaceskip=\fontdimen7\font\relax
    \ttfamily
    #1%
  \endgroup
}

\supsubsection{Introduction}
LIFEPLAN (\href{https://www.helsinki.fi/en/projects/lifeplan}{https://www.helsinki.fi/en/projects/lifeplan}) is a six-year initiative funded by the European Research Council (ERC) to study biodiversity in a worldwide sampling program over multiple years, using different methods to gather data across a broad range of taxonomic groups. Methods include, for example, camera traps and sound recordings. Arthropod communities are sampled using Malaise traps, a commonly used insect trap where organisms are collected in bulk and preserved in ethanol. The identification of species collected in these traps is primarily done using a technique known as DNA metabarcoding where a short gene fragment is extracted and sequenced simultaneously for all specimens in the sample. Standardized photographs of the samples spread across a white surface are also taken and provide additional information. Further description of the arthropod data is available in \flexiref{section:d_descr}{Data description}. The objective of this project within LIFEPLAN is to develop a machine learning algorithm to classify and count arthropods from images of bulk arthropod samples. 

A reliable classification algorithm for images of bulk arthropod samples has the potential to accelerate and automate insect biodiversity surveys without the need for expensive analyses. Further, it is challenging to obtain abundance estimates from metabarcoding data alone. Thus, abundance estimates and counts from images can be used as a complement or verification of estimates based on molecular data. 

\supsubsection{Getting started} 
\supsubsubsection{Some useful terminology}
To annotate an image means adding labels to it. This can be done in many different ways (for example, using points, see Figure~\ref{fig:annot} left). We have chosen to use segmentation masks, which means that we draw a contour around each object (arthropod, or `bug') we want to label (Figure \ref{fig:annot} right). When correctly annotated, each pixel that is part of that bug should be inside the mask. 

\begin{figure}[h!]
    \centering
    \includegraphics[width=0.4\linewidth]{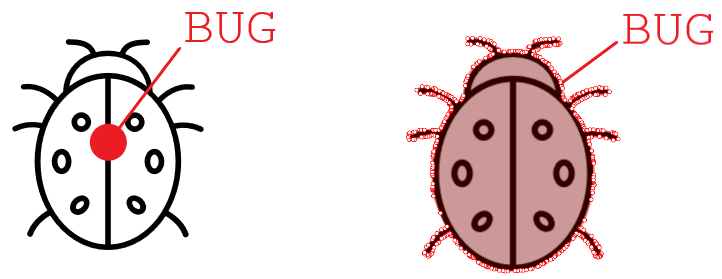}
    \caption{Annotation using points (left) and segmentation masks (right).}
    \label{fig:annot}
\end{figure}

In the program we use for annotation, the different objects are called entities. Each bug should be a separate entity, which implies that each entity should contain only one segmentation mask. 

Creating all the segmentation masks from scratch would take a very long time, so we use a simple method known as a watershed algorithm to generate a first draft. A watershed algorithm finds and outlines dark areas against a white background, or in our case, the different arthropods. The masks, however, are not perfect, and most need some degree of correction (Figure~\ref{fig:segmentation}). The annotation task can thus be divided into two main steps: (1)~create and correct segmentation masks and (2)~assign taxonomic labels to the segmentation masks. The first task requires the annotator to have basic knowledge of arthropod morphology (in this document referred to as `non-expert'), while the second task requires detailed taxonomic knowledge (here referred to as `expert').

\begin{figure}[h!]
    \centering
    \refstepcounter{suppfigure}
    \includegraphics[width=0.5\linewidth]{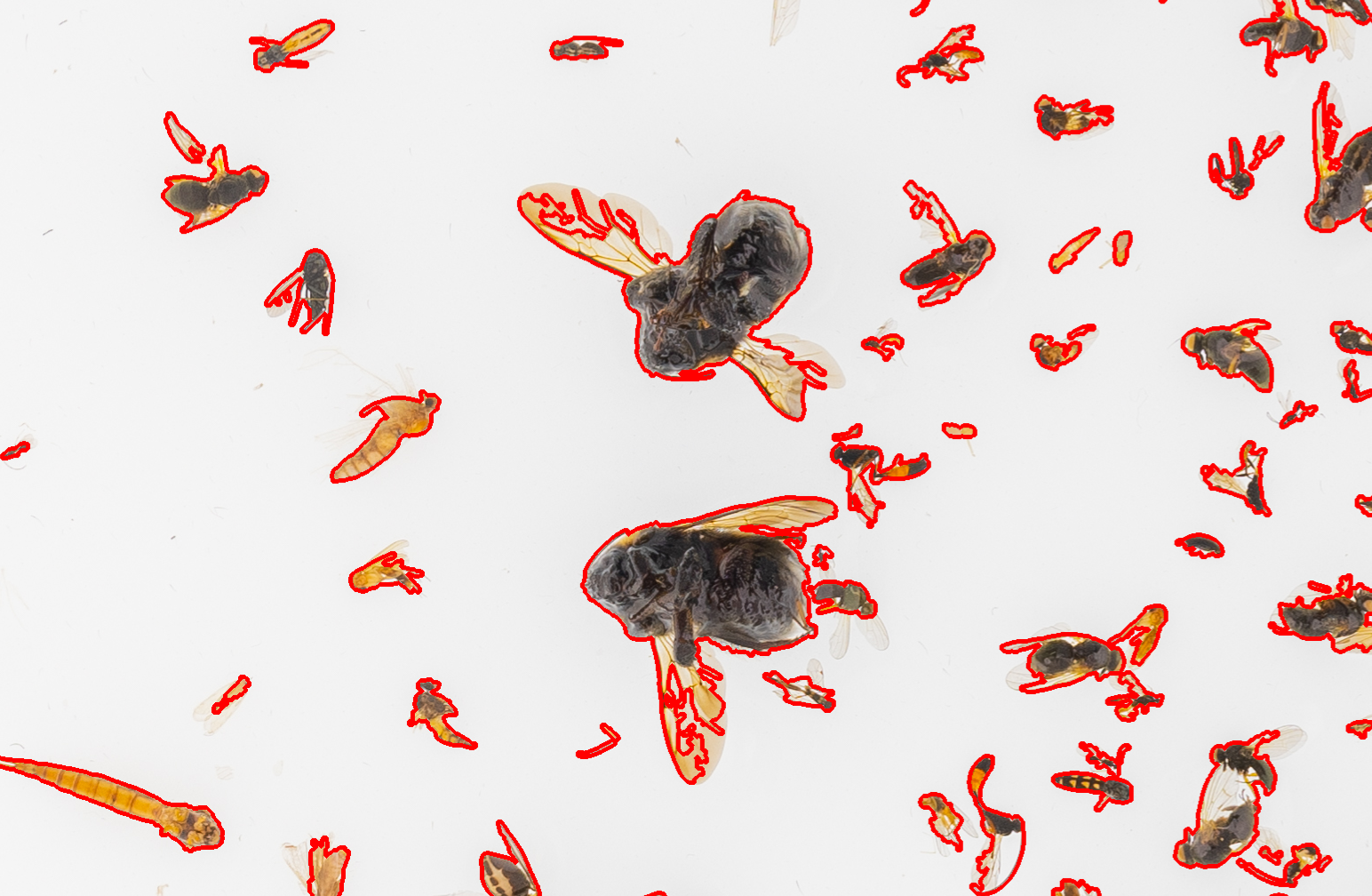}
    \caption{Initial segmentation masks generated by a watershed algorithm.}
    \label{fig:segmentation}
\end{figure}

\supsubsubsection{Data description} \label{section:d_descr} 
There are four types of data relevant for this project, generated by either bulk analysis of samples (DNA metabarcoding data and bulk images) or analysis of individual specimens from the same samples (individual DNA barcoding data and individual images, here also known as `Keyence images'). 

In general, in the wider Lifeplan project, Malaise trap samples are treated in ``bulk'', i.e. DNA is extracted from the full sample (all individuals together) at once. The resulting broth is then sequenced for a diagnostic fragment of the COI gene region; a process called metabarcoding. After metabarcoding, samples are also photographed in bulk, i.e. all individuals in one image. For this purpose, the content of the bottle is spread out on a white tray, in an effort to separate the individual bugs as well as isolate the debris. Depending on the quantity of arthropods in the sample, multiple trays may be needed, resulting in more than one bulk image per sample. As a result of this workflow, each bulk image (i.e., a picture of tens to thousands of arthropod individuals) is associated with a list of DNA-based taxonomic units (``species''). Nonetheless, since the sample is treated as one entity, individual entries within the list cannot be associated with individual arthropods in the photograph. 

In this project, we are working with a subset of 45 samples where all arthropod specimens, in addition to the bulk analyses, were also individually analyzed. The individual arthropods were picked from the original bulk samples and placed in separate wells in a 96-well plate. For arthropods that were too large for wells, a leg was extracted from the specimen and placed in the well. Each specimen was then individually barcoded, i.e. the barcode region of the COI gene is sequenced. This is largely the same region as is targeted in metabarcoding, but it is possible to analyze slightly longer sequences (around 650 base pairs or positions, as compared to around 420 base pairs in metabarcoding). After molecular analyses, the arthropods were individually photographed (these images are also referred to as Keyence images). Arthropods that were too large to fit in the wells were pinned and photographed separately. Contrary to the bulk analysis, this procedure results in a set of specimens with a direct association between the DNA identification and an image.

\begin{figure}[b!]
    \refstepcounter{suppfigure}
    \centering
    \includegraphics[width=0.75\linewidth]{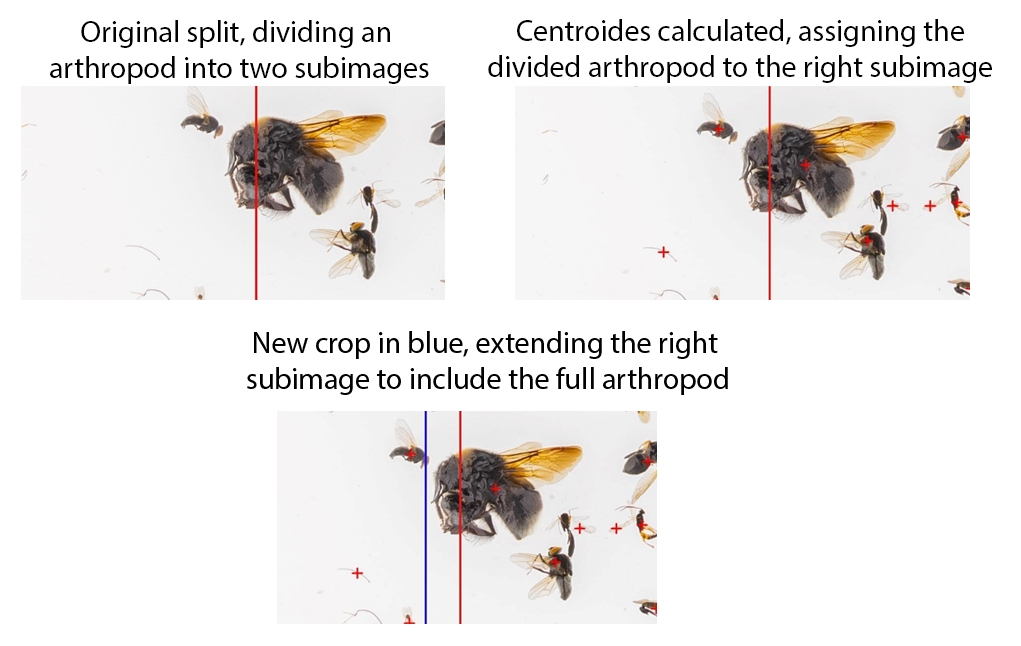}
    \caption{When images were split into subimages for annotation, the exact location of the split depended on the location of arthropods, to ensure that arthropods were not divided into multiple subimages.}
    \label{fig:crop_adj}
\end{figure}

Arthropod samples can contain around \num{3000} individual arthropods. To enable fast rendering of segmentation masks in TORAS, we have chosen to split images into $4 \times 4$ subimages for annotation. To ensure that no arthropods were split between two subimages, we used the initial watershed masks to indicate the location of the arthropods in the images. We calculated the centroid of each mask and assigned the mask to the subimage containing the centroid. We then adjusted the size of each subimage to include the full range of each segmentation mask assigned to that subimage, plus a buffer of 100 pixels to allow for extending the mask in any direction during manual editing (see Figure \ref{fig:crop_adj}). This method of splitting images resulted in some overlap between subimages, where arthropods appeared in more than one subimage. To reduce the risk of annotators marking the same arthropod twice in two different subimages, arthropods that appeared in a different subimage than where they were originally assigned were marked by displaying the segmentation mask in the subimage.

\supsubsubsection{The Toronto Annotation Suite} 
The annotation is done in the Toronto Annotation Suite (TORAS), a web-based annotation platform. Create a free account and log in here: \href{https://aidemos.cs.toronto.edu/toras/login}{https://aidemos.cs.toronto.edu/toras/login}. Send your username to the administrators of the project, and they will add you to the correct project on TORAS. 

To get started with annotations, choose \texttt{GO TO YOUR TASKS PAGE}.

\includegraphics[width=0.6\linewidth]{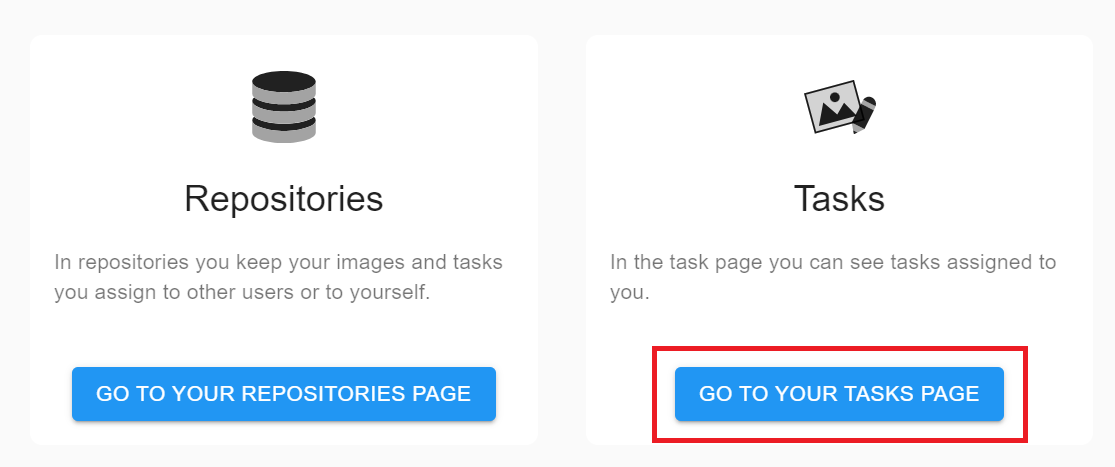}

There, you can see all your assigned tasks (that is, images to annotate) and their status. Click \texttt{SELECT} to get started with a task.

\includegraphics[width=.95\linewidth]{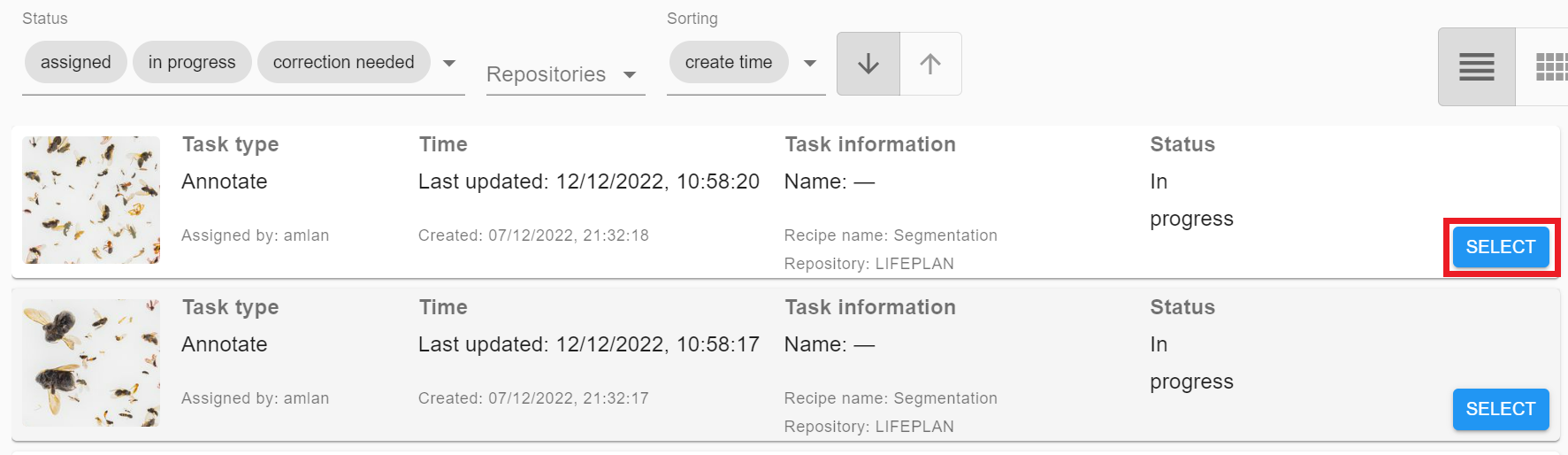}

This is what the annotation page looks like: 

\includegraphics[width=.85\linewidth]{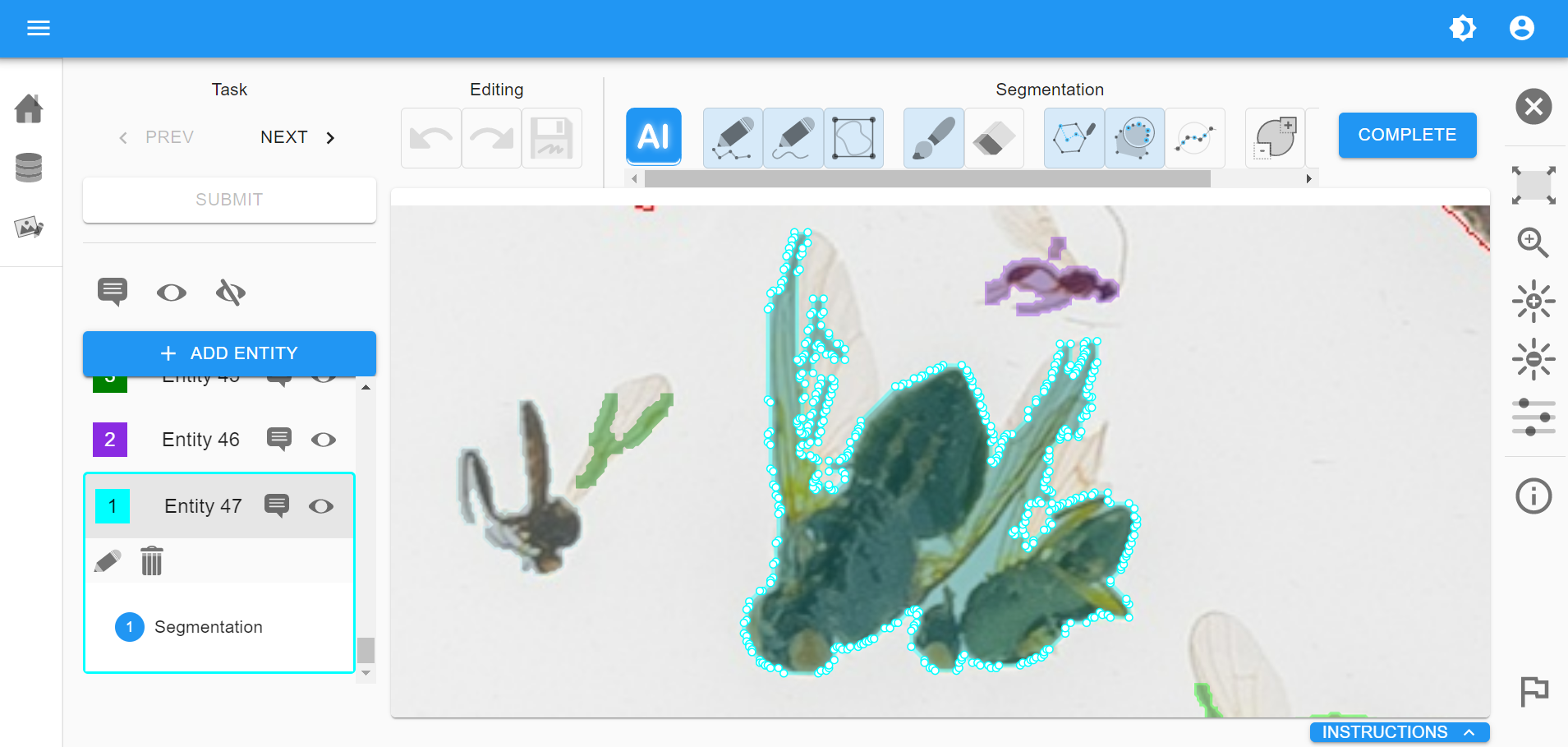}

In the right panel, you have different options for zooming in/out, as well as changing brightness, mask opacity, etc. In the left panel, there is a list of the different entities in the image, each with one segmentation mask. You can zoom in on a segmentation by clicking on the entity name.

For a brief tutorial on how to use the different segmentation tools, visible in the top panel, please see \href{https://www.youtube.com/watch?v=jrr0N4zpriw}{this video}.

\supsubsection{Create masks + base annotations (non-expert)}

\supsubsubsection{Correcting masks}

When images are uploaded to TORAS, they have associated masks that have been created with the watershed algorithm. The masks are automatically refined with a TORAS algorithm, but still need manual adjustments (for example, to include legs, or to remove background from the mask) and quality checks:

\begin{enumerate}
    \item For each polygon mask:
    \begin{enumerate}[label=\alph*.]
        \item Change entity name (shortcut: \texttt{n}) to one of these short names:

        \begin{center}
        \begin{tabular}{>{\centering\arraybackslash}m{2cm}>{\centering\arraybackslash}m{2cm}m{10.5cm}}
            \toprule
            \textbf{Short name} & \textbf{Meaning} & \textbf{Explanation} \\
            \midrule
            b & bug & Any arthropod \\
            \midrule
            u & unknown & Could be an arthropod, but I can’t tell from the image \\
            \midrule
            d & debris & Any debris, including e.g. loose legs, wings, etc. \\
            \midrule
            e & edge & Mask containing tray edge, QR code, etc. I.e. not debris, not an arthropod \\
            \bottomrule
        \end{tabular}
        \end{center}

        \item For each mask identified as \texttt{b}, correct the mask:

        \begin{enumerate}[label=\roman*.]
            \item Include all body parts, such as legs, antennae, and wings (e.g. using the painter, shortcut: \texttt{r}). The TORAS algorithm is also good at estimating segmentation masks, so if the mask is very poor, it is sometimes faster to delete the mask and draw a new one using the bounding box tool (shortcut: \texttt{b}).

            \item Exclude areas with only background, e.g. between legs (e.g. using the eraser, shortcut: \texttt{e}). See Figure \ref{fig:gradient_instruction}: in the left panel, too much background is included. In the middle panel, the background between the legs has been removed while the legs themselves are still kept inside the mask.

            \textit{Exception:} if a bug has more than eight legs, or is very hairy, you don’t have to exclude the background between the legs/hairs. However, you should still make sure that all parts of the bug (legs and hair included) are inside the mask (see Figure \ref{fig:many_legs} for an example).

            \item Mask should be relatively snug. Especially for small bugs or low-quality images, the edge of a bug can be blurred when zoomed in, creating a gradient between the bug (dark) and the background (light). When adjusting the mask, the edges of the mask should run approximately in the middle of such a gradient (see Figure \ref{fig:gradient_instruction}).

            \textit{Tip for facilitating adjustments}: if the density of points along the mask is very high, it might be easier to make adjustments if you first reduce the density (see Figure \ref{fig:high_density}).

            \item Mark entity as complete (shortcut: \texttt{c}).

            We are annotating a large number of arthropods, and so each annotation should not take a lot of time. The rule of thumb is to spend no more than 10 seconds per bug mask once you are familiar with the tools and procedure. Use this as a guideline for how detailed you should be when adjusting the masks.
        \end{enumerate}


\begin{figure}[h!]
    \refstepcounter{suppfigure}
    \centering
    \includegraphics[width=0.75\linewidth]{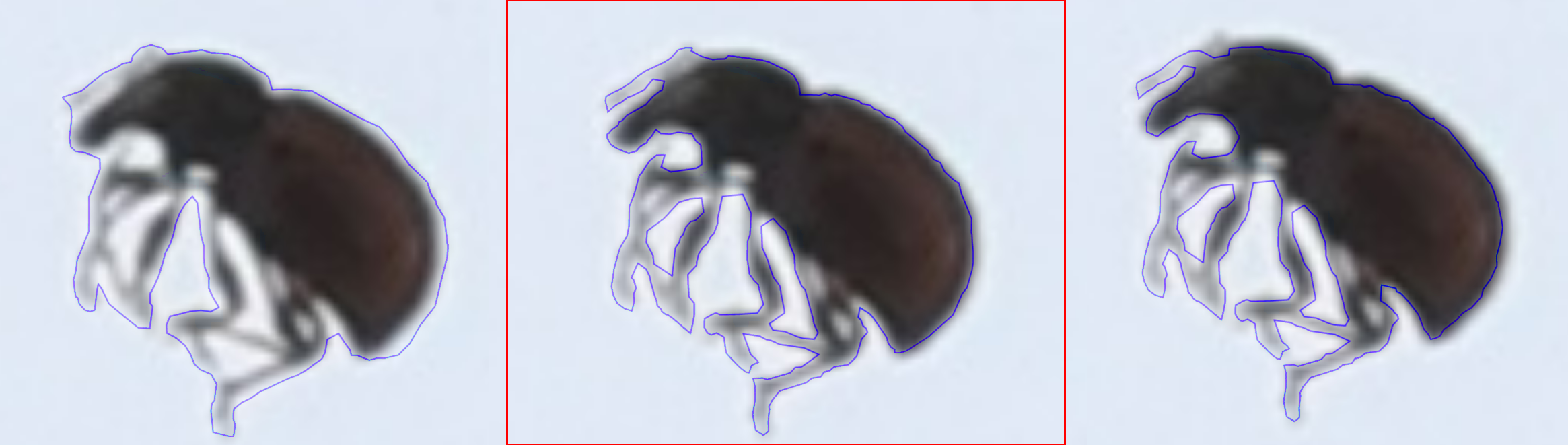}
    \caption{The edge of a bug can be blurred when zoomed in, creating a gradient between the bug (dark) and the background (light). When adjusting the mask, the edges of the mask should run approximately in the middle of such a gradient. Focusing on how the mask looks on the back of the insect in the picture: to the left, the mask is a bit too big, in the middle it looks good, and to the right, the mask is a bit too small. The middle and right image also shows how areas of background between the legs should be removed.}
    \label{fig:gradient_instruction}
\end{figure}

\begin{figure}[h!]
    \refstepcounter{suppfigure}
    \centering
    \includegraphics[width=0.7\linewidth]{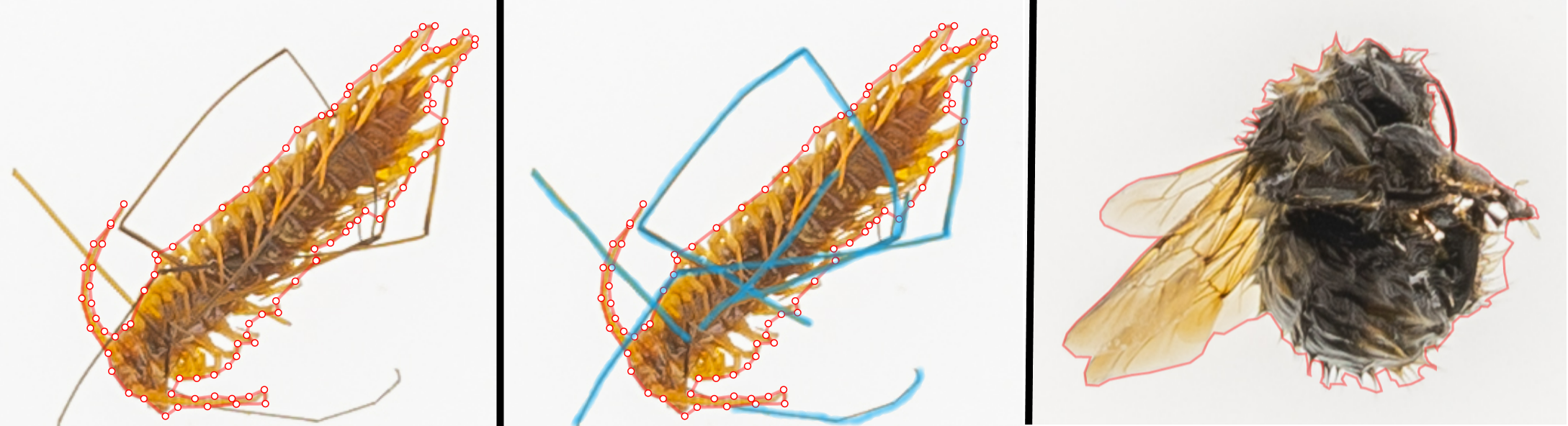}
    \caption{Example of acceptable masks for bugs that either have more than eight legs or are very hairy. All parts of the bug should be inside the mask, but not all areas of background between legs/chunks of hair must necessarily be removed. Which bugs are too hairy to make exact masks is a judgment call, but use the rule of 10 seconds per bug to guide you. The many-legged bug in this example is particularly tricky due to the debris tangled among the legs (marked with blue colour in the middle image).}
    \label{fig:many_legs}
\end{figure}

\begin{figure}[h!]
    \refstepcounter{suppfigure}
    \centering
    \includegraphics[width=0.75\linewidth]{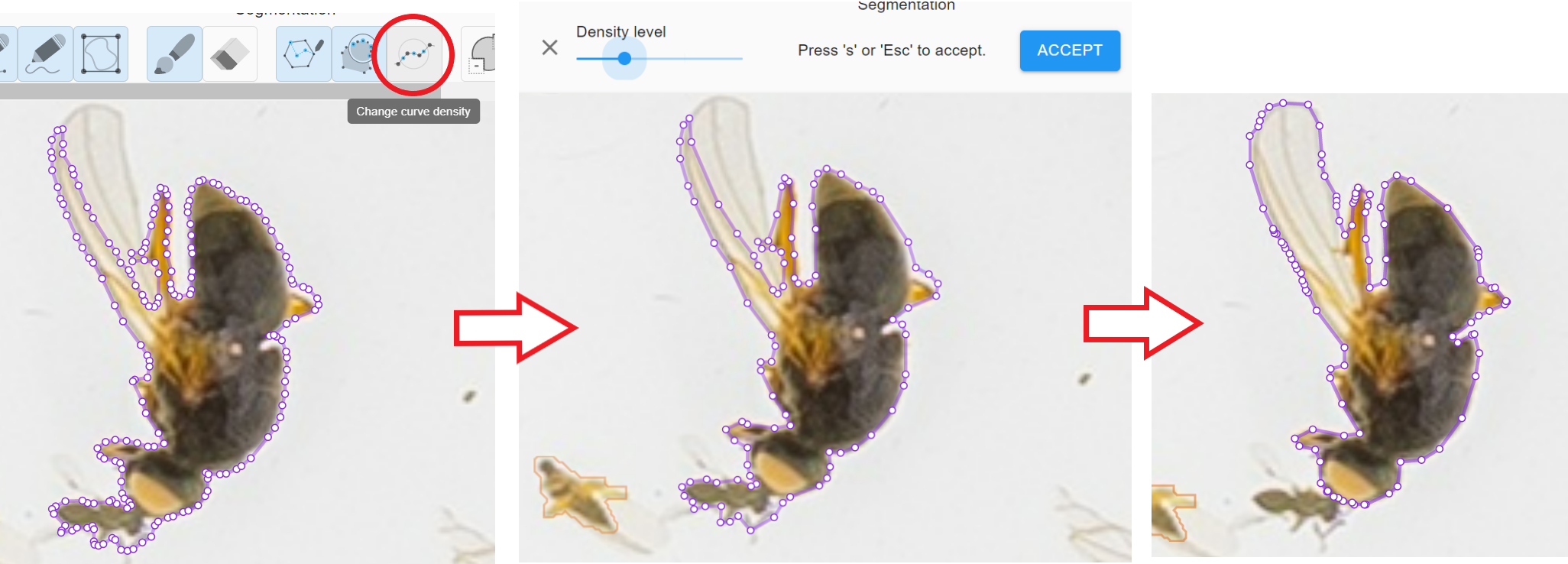}
    \caption{Reducing density of points along a mask. Use the tool ``Change curve density'', and click on two points to mark the section you wish to edit (click on the same point twice to edit point density along the entire mask). Slide the scale to the density you want. Press Esc to accept the changes. To the right, you can see the finalized mask.}
    \label{fig:high_density}
\end{figure}

    \item If multiple bugs have been grouped together in the same mask, or if one bug is superimposed on another: 

\begin{enumerate}[label=\roman*.]
            \item Select one bug to start with and adjust the mask for that bug, as described in 1b (see Figure \ref{fig:step7a}).

            \item Add a new entity (shortcut: \texttt{+}). Rename it (shortcut: \texttt{n}) to `b' (bug). 

            \item Create a mask around the next bug, for example, using the bounding box tool.

            \item Adjust the mask as described in 1b.

            \item Repeat steps ii-iv for each unmasked bug.

           If you can’t distinguish which bug a certain body part belongs to (e.g. if the legs are tangled), leave the body part in question out of the mask.
        \end{enumerate}

\begin{figure}[h!]
    \refstepcounter{suppfigure}
    \centering
    \includegraphics[width=0.7\linewidth]{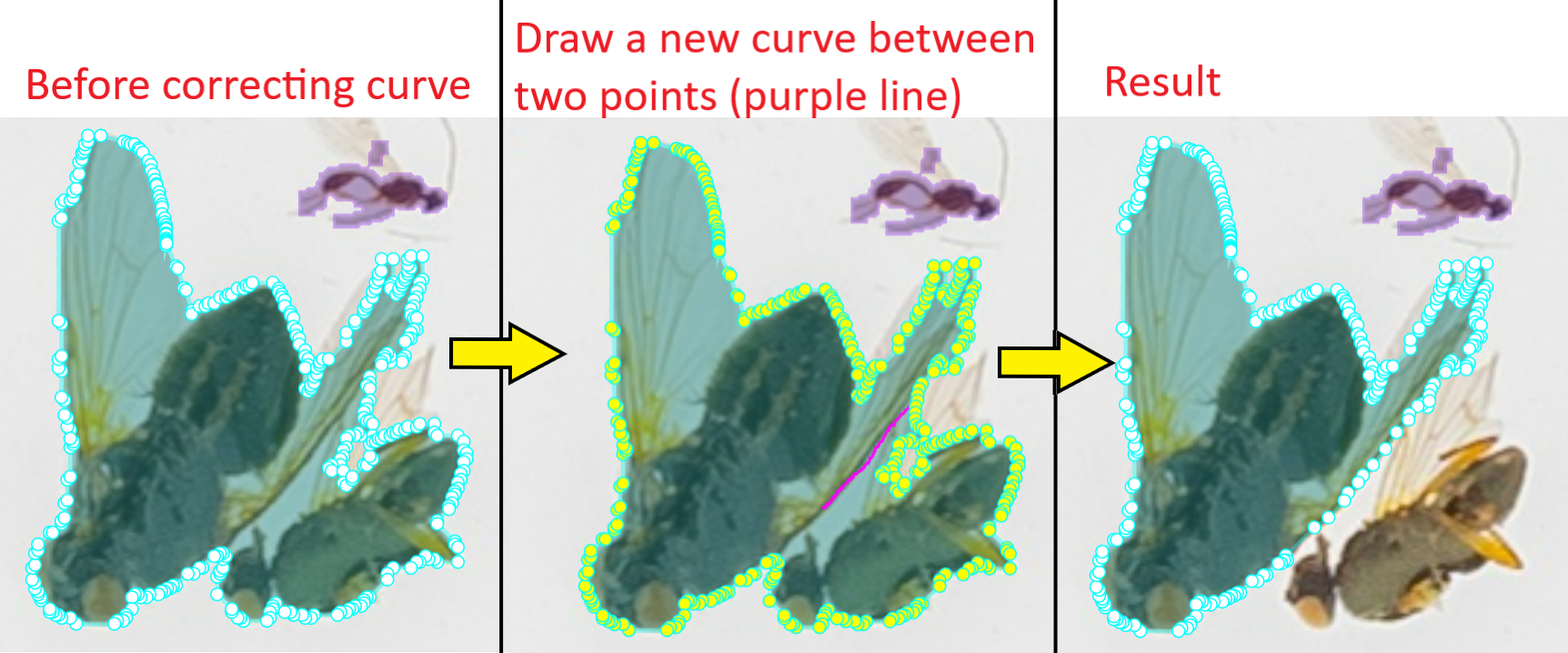}
    \caption{An example of two insects grouped within the same mask. Here, `Correct part of curve' (shortcut: \texttt{s}) is used to draw a new edge to exclude one of the insects. The next step would be to create a new entity and draw a mask for the excluded insect.}
    \label{fig:step7a}
\end{figure}

\item For masks that are not bugs (i.e., ``u'', ``d'', or ``e''), mark the mask as complete without making any adjustments.

    \end{enumerate}
    \item Look through the image for arthropods missed by the watershed algorithm. For each unmasked bug that is fully contained in the image and not outlined with red colour (Figure \ref{fig:red_outline}): 

\begin{enumerate}[label=\alph*.]
\item Add a new entity and mask as described in 1c ii-iii.
\item Adjust the mask as described in 1b.

\begin{figure}[h!]
    \refstepcounter{suppfigure}
    \centering
    \includegraphics[width=0.6\linewidth]{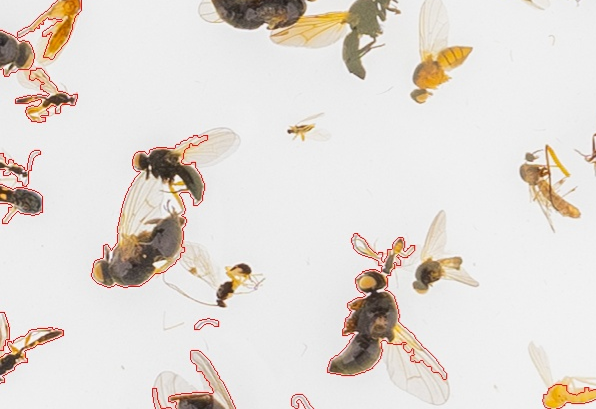}
    \caption{If the bugs are outlined with a red colour, it means that they are duplicated from another image, and you should not create a mask around them. The outlined bugs are found near the edges of the image.}
    \label{fig:red_outline}
\end{figure}

\end{enumerate}
    \item When you are done, mark all entities as complete, and click ``Submit''.
\end{enumerate}

\supsubsubsection{Tips and tricks}

\begin{itemize}
    \item Make use of the keyboard shortcuts in TORAS. They can be displayed by pressing ``?'' on your keyboard. Here are some examples: 
\begin{table}[h!]
\centering
\begin{tabular}{lc}
\toprule
\textbf{Action} & \textbf{Keyboard Shortcut} \\
\midrule
Add entity & \texttt{+} \\
Paint & \texttt{r} \\
Erase & \texttt{e} \\
Rename entity & \texttt{n} \\
Create mask using bounding box & \texttt{b} \\
Complete entity & \texttt{c} \\
\bottomrule
\end{tabular}
\end{table}
    \item Sometimes when a new entity is created, TORAS zooms out to show the full image. This is optimized for annotation of larger objects, but is not ideal when having to relocate a tiny bug in our images. To change this behaviour, click on options (right panel; \includegraphics[width=0.04\linewidth]{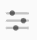}) and make sure that `Auto zoom on action' is turned off. 

    \includegraphics[width=0.35\linewidth]{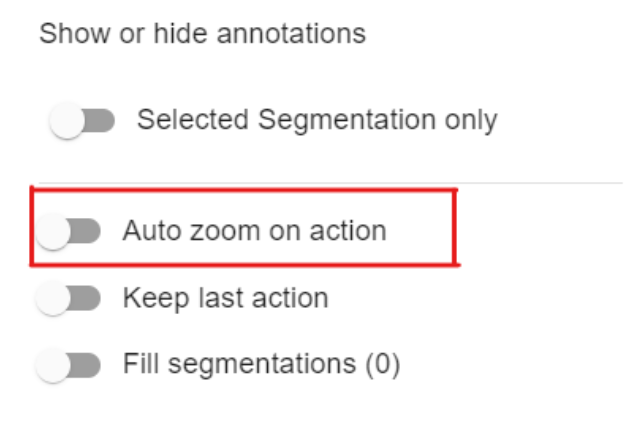}
\end{itemize}

\supsubsection{Create labels (expert)}
\label{sec:expert_instructions}

\supsubsubsection{The task} 
All bugs in the image should now have a refined segmentation mask, and the main task of the expert annotator is to assign a taxonomic label to each bug mask. As described previously, in addition to the images, we have data from DNA (meta-)barcoding of each sample; this sample-specific taxonomic information will be used to delimit the taxonomy from which you choose the labels, and might thus work as a guidance when annotating the images. 

DNA barcoding results in classification on varying taxonomic levels (mainly due to variable taxonomic cover of reference databases); some arthropods are identified to species, while others are only identified to order. It is also possible that specimens have been misidentified using the genetic data, for example if two genera have overlapping genetic variation, or if the reference database contains misidentified sequences. Therefore, you are free to choose taxonomic labels that are not part of the sample-specific suggested taxonomy (see instructions below).

\textit{NB: each original image is divided into sub-images before they are uploaded to TORAS (see previous description), and the sample-specific DNA data thus corresponds to multiple sub-images. Further, the original sample might have been split into two full images because of a large quantity of insects, in which case the number of corresponding sub-images is of course even higher.} 

Before labeling a bug, you are asked to validate the mask created by the non-expert annotator. This is mainly to catch any major mistakes, and you are not generally expected to adjust details of the mask (specific instructions are given below). 

\supsubsubsection{Validate masks}
For entities marked as `b' (bug):  

\begin{enumerate}
    \item Check the mask of the non-expert annotator. 
    \item If it looks ok, go to classification. 
    \item If there are big mistakes (see examples below), first correct the mask (for how to do this, see below and consult the instructions above for the non-expert annotator), then go to classification. 
    
    \textit{Examples of big mistakes:}
    \begin{itemize}
        \item a visual characteristic important for classification of the bug is not included in the mask (e.g., a wing, antennae)
        \item a whole body part of the bug is not included in the mask
        \item two bugs are grouped within the same mask 
        \item the mask includes a considerable amount of debris
    \end{itemize}
\end{enumerate}

To enable editing of the mask, first exit the classification pane by clicking anywhere outside of it. Then click on \texttt{Segmentation} of the entity you want to correct in the left pane: 

\includegraphics[width=0.25\linewidth]{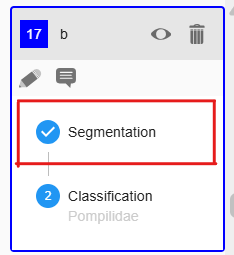}

Then click on \includegraphics[width=0.15\linewidth]{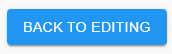} in the top right corner.

\supsubsubsection{Classification}
When you choose an entity in the list on the left side of the screen, the classification window opens:

\includegraphics[width=0.95\linewidth]{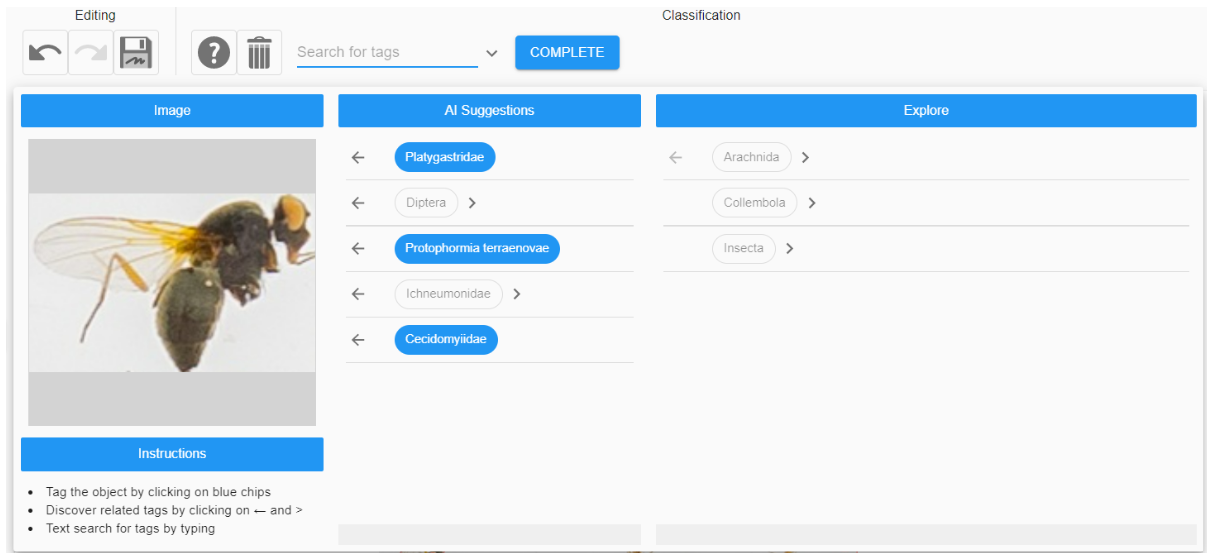}

To the left, you see an image of the bug, and to the right, you can click through the sample-specific taxonomy to find an appropriate label. It is also possible to search for labels, using the search field above the classification window (`Search for tags'). \textit{In the middle of the classification window, there are some AI-suggested labels -- these are unlikely to be correct (the tool is meant to work for a wide variety of objects, and is not specialized on arthropods).} 

If you can’t find the label you are looking for in the sample-specific taxonomy, you are free to add another label manually. You do this by writing the label in the search field and then pressing \texttt{ENTER}. We have prepared two excel sheets containing all insect taxa in Sweden (where the majority of samples are from): one going down to family level (\texttt{dyntaxa\_boldified\_family.xlsx}) and one to species level (\texttt{dyntaxa\_boldified\_species.xlsx}). To avoid misspellings of manually added labels, please copy the taxonomic label from those files. 

When you have chosen a label, press `Complete' to mark the entity as done.

Now for the million-dollar question: \textbf{``How confident should I be of the classification?''} First of all, because of the large number of bugs to classify, you should not spend more than on average \textbf{10 seconds} on the classification of a single bug, so that puts some limitations on how detailed you can be. However, it will likely often be the image quality and lack of detail that limits the taxonomic level you can get to. In general, you should assign the most detailed taxonomic label you feel confident is correct. To capture more detailed information, it is also possible to give additional labels on a lower confidence level. There are two ways to do this: 

\begin{enumerate}
    \item When you want to choose \textbf{one} label: include a label on a higher taxonomic level. The label on the highest taxonomic level will then be interpreted as `high confidence', while lower levels are interpreted as `low confidence'. 
    
    \textit{Example:} you think the bug you are looking at belongs to the family Ichneumonidae, but you could be mistaken. Therefore, you add both ``\texttt{Ichneumonidae}'' and ``\texttt{Hymenoptera}'' as labels. Hymenoptera will then be interpreted as `high confidence', Ichneumonidae as `low confidence'.
    \item When you want to choose \textbf{multiple} labels: choose multiple labels on any taxonomic level. The last common ancestor of all your chosen labels will be interpreted as `high confidence', whether or not you include it as a label. 
    
    \textit{Example:} 
    
\includegraphics[width=0.3\linewidth]{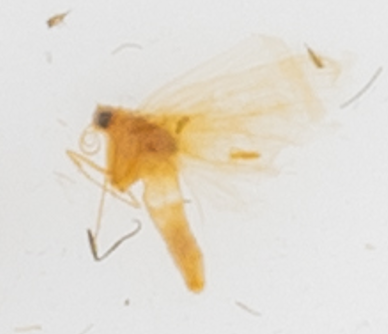}

    You think this bug belongs either to Lepidoptera: Tortricidae or Lepidoptera: Geometridae. Therefore, you choose ``\texttt{Tortricidae}”'' and ``\texttt{Geometridae}'' as labels. They are both interpreted as `low confidence' labels, but as they both belong to the order Lepidoptera, ``\texttt{Lepidoptera}'' is interpreted as a `high confidence' label.

    \includegraphics[width=0.2\linewidth]{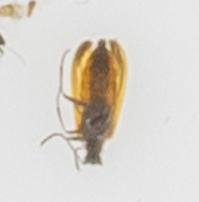}

    Similarly, you think this is either a ``\texttt{Coleoptera}'' or ``\texttt{Hemiptera}'', which are interpreted as `low confidence' labels. They both belong to the class Insecta, so ``\texttt{Insecta}'' is interpreted as a `high confidence' label.
\end{enumerate}

As long as you give a single label to a bug, it will be interpreted as `high confidence'. 

\supsubsubsection{Images of individual bugs}
For each bulk image, we also have the full set of bugs photographed individually (the so-called ``Keyence images''). These images can be used as guidance, for example, to get an idea of the abundance and morphological distribution of certain families. Images are grouped by order and family and can be accessed locally from an external hard drive. 

\clearpage
\supsection{Supplementary methods} \label{sec:supplementary_methods}

\supsubsection{Determining upsampling factor for tiles}
\label{sec:upsampling-factor}
If an entire bulk insect sample is downsampled to fit within a model's input size of $1024 \times 1024$ pixels, each insect is rendered at a lower resolution than in the original image, leading to blurred contours and fewer visible details -- especially problematic for detecting small insects. An alternative is to divide the images into tiles to preserve visual details. Using smaller tiles than the required input size and instead upsampling the images to target resolution can affect model performance. For example, presenting images to models at higher resolutions allows the model to spend more compute in processing the full input image, potentially improving its performance \citep{efficientnet}. Correspondingly, we investigated how much the model's performance could be increased if the original images were upsampled before presenting them to the model. 

To determine the optimal upsampling factor for our instance segmentation models, we performed training and inference while varying the dimensions of the bulk image tiles. As we decreased the size of our tiles, we needed to increase the upsampling rate to reach our fixed input size of $1024 \times 1024$ pixels. We performed this analysis on the validation set using the SAHI approach to ensure this hyperparameter selection was not based on the test partition. For each trial, we maintained a fixed input size of $1024 \times 1024$ pixels, a common input size for pretraining instance segmentation models \citep{he_mask_2018, li_mask_2022, cheng_masked-attention_2022}. Tiles smaller than this input size were upsampled to $1024 \times 1024$ pixels using bilinear interpolation. Thus, a tile size of $1024 \times 1024$ pixels would require a zoom factor of one to reach our desired input size, $512 \times 512$ pixels would require a zoom factor of two, and so forth until $128 \times 128$ pixels, which would require a zoom factor of eight.

As the zoom factor increases, the relative size of the arthropods in each tile increases, although each tile includes less spatial context, and more arthropods are cut between tiles. We observed that all three models achieve the best mask AP when they use $512 \times 512$ pixel tiles or a zoom factor of two (Figure \ref{fig:tile-analysis}), which we consequently used for all further experiments. 

While increasing the zoom factor from one to two improves instance segmentation performance, higher zoom factors gradually degrade performance. Very small tiles, with zoom factors of six and eight, showed the worst mask AP across all models, suggesting that the increase in relative size is offset by the lack of spatial context when small tiles are used. Such context may be important for distinguishing small insects from surrounding debris. For example, as the tile size decreases, more insects are split between tiles. These partial insects may be more difficult to distinguish from debris, which includes loose insect legs and wings. This also complicates the inference stage as a) our models must correctly identify partial insects, and b) the SAHI algorithm must correctly merge fragmented insect predictions across tiles.

\begin{figure}[h!]
    \centering
    \includegraphics[width=0.7\linewidth]{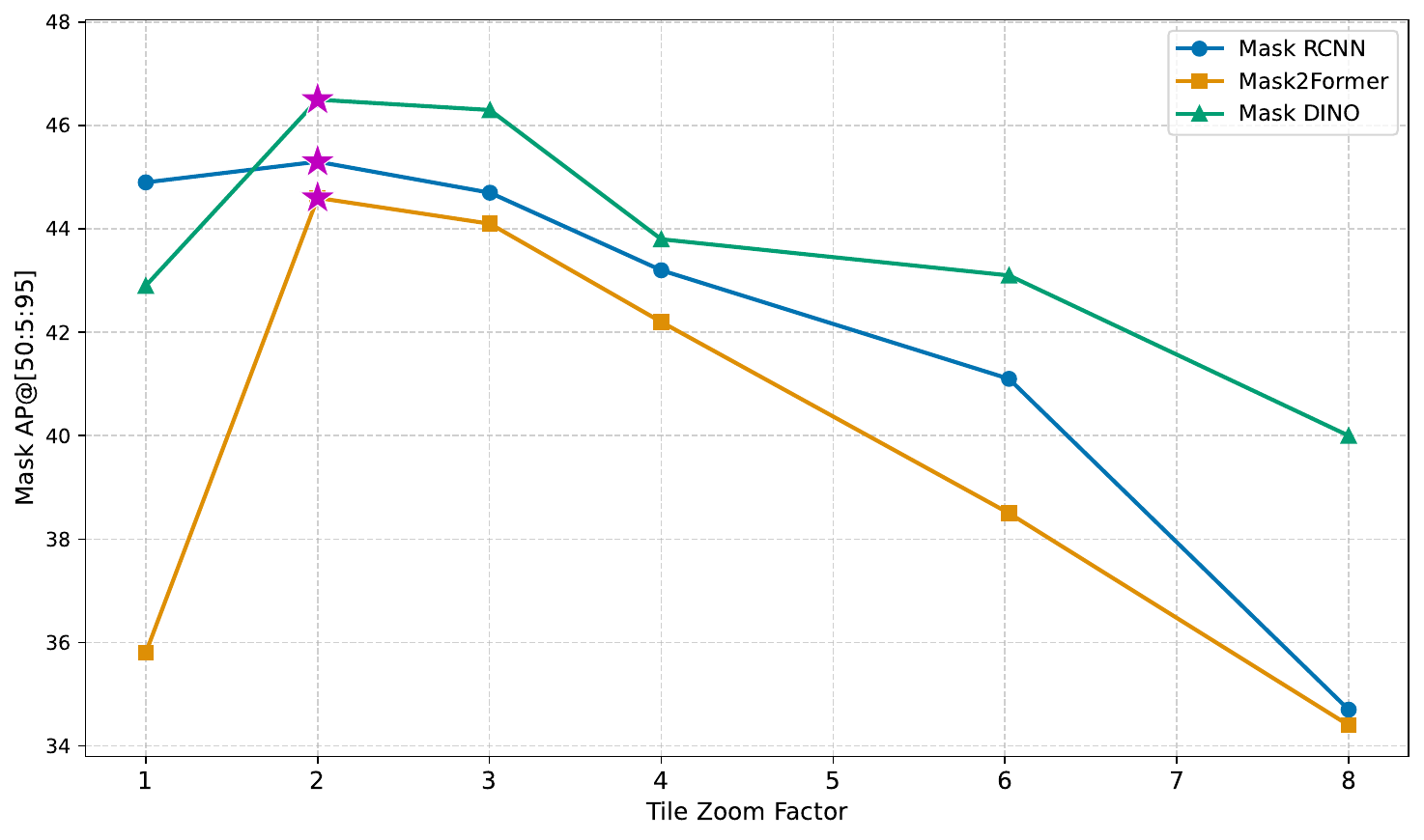}
    \caption{Validation mask AP versus tile zoom factor for our supervised baselines: Mask R-CNN, Mask2Former, and Mask DINO. For all three models, a zoom factor of two, corresponding to a tile size of $512 \times 512$ pixels, which is upsampled to a $1024 \times 1024$ pixels model input, gives the best instance segmentation performance.}
    \label{fig:tile-analysis}
\end{figure}

\supsubsection{Implementation details for zero-shot methods}
\label{sec:zero-shot}
First, we assessed the Cut and Learn (CutLER) model \citep{wang_cut_2023}, an unsupervised instance segmentation method trained on a dataset without human annotations. CutLER leverages a self-training process where the model is iteratively trained on its own predictions to refine the quality of subsequent instance masks. For all CutLER experiments, we used a self-trained Cascade Mask R-CNN checkpoint (\opus{cutler\_cascade\_final}) \citep{wang_cut_2023}. 

We then evaluated Grounding DINO \citep{ren2024grounded, liu2023grounding} and Florence-2 \citep{xiao2023florence}, which can localize objects of interest through text prompts. These text prompts can denote simple category names or referring expressions. For all Grounding DINO experiments, we provided the Grounding DINO-B model (\opus{groundingdino\_swinb\_cogcoor}) \citep{liu2023grounding} with the prompt \texttt{insect.}, where ``.'' is used as a delimiter for different object classes. We then used the default box and text thresholds of 0.35 and 0.25, respectively.

For our Florence-2 evaluations, we used the publicly available \texttt{Florence-2-large-ft} checkpoint \citep{xiao2023florence}. In addition to a text prompt, Florence-2 requires a task prompt denoting whether to perform captioning, detection, or other vision-language tasks. 
 
Thus, we provided the following prompt to Florence-2: \opus{<OPEN\_VOCABULARY\_DETECTION> small brown-yellow insects}. To suppress large bounding box predictions, we filtered out bounding boxes that occupy more than 40\% of the area of a given $512 \times 512$ pixel tile (for comparison, the largest specimen in the dataset had a ground truth mask equal to 32\% of a tile). 

Lastly, we leveraged Gemini 2.0 Flash’s spatial understanding capabilities to perform object detection \citep{gemini_flash_2}. With a temperature of 0.5, we provided the following system instructions: \opus{Return bounding boxes as a JSON array with labels. Never return masks or code fencing. Limit to 50 objects. Never repeat or duplicate bounding boxes. If an object is present multiple times, return the same label for each instance.}

When performing detection, we used the following text prompt: \opus{Detect the 2d bounding boxes of the small brown insects, ants, flies, and/or gnats. Exclude loose wings, legs, and debris.} As with Florence-2, bounding boxes occupying more than 40\% of a tile were filtered out before being used as prompts for SAM 2.1. We performed inference with the \opus{sam2.1\_hiera\_large} checkpoint without any fine-tuning on the MassID45 training set \citep{ravi2024sam2segmentimages}.

\supsubsection{Model evaluation with tailored confidence thresholds}
\label{sec:confidence-thresholds}
To determine appropriate confidence thresholds for each model, we plotted precision-recall (PR) curves using their predictions on the MassID45 validation set (see Figure \ref{fig:pr}). We fixed the IoU threshold to 50\%, indicating that we consider predicted masks as correct if they overlap by more than 50\% in area with the ground truth. Each point corresponds to the precision and recall at a particular confidence threshold. Thus, for each model we selected the confidence threshold with the highest F1-score -- the harmonic mean between precision and recall. This optimal confidence threshold is generally the point closest to the top-right corner of the PR curve, which represents perfect precision and perfect recall. These confidence thresholds can be interpreted as suggested operating points for each model when used on bulk images in a real-world setting. 

Using these tuned confidence thresholds, we performed inference on the MassID45 test set, then filtered out any predictions below each model’s confidence threshold. We then measured the number of TP, FP, and FN pixels predicted by each model on the test set (see Table \ref{tab:segmentation-areas}). Consistent with our exemplar patch in Figure \ref{fig:example-detections}, Mask DINO predicts the highest number of TP pixels and lowest number of FN pixels, while Grounding DINO has the highest proportions of FPs and FNs. Mask R-CNN predicts the fewest FPs, while Mask2Former generally achieves a balance between Mask DINO and Mask R-CNN.

\begin{table}[h]
\caption{Proportion of TP, FP, and FN pixels for each model on the MassID45 test set after tuning confidence thresholds.}
\label{tab:segmentation-areas}
\centering
\begin{tabular}{lrrr}
\toprule
\textbf{Model} & \textbf{TP Area} & \textbf{FP Area} & \textbf{FN Area} \\
\midrule
Grounded SAM 2.1 & \num{712478} (63.9\%) & \num{253701} (22.8~\%) & \num{148456} (13.3~\%) \\
Mask2Former      & \num{783319} (80.7\%) & \num{110204} (11.4~\%) & \num{77615} (~7.99\%) \\
Mask DINO        & \textbf{\num{787067}} (80.7\%)& \num{114215} (11.7~\%) & \textbf{\num{73867} (~7.57\%)} \\
Mask R-CNN       & \num{777120} (81.4\%)& \textbf{\num{ 93473}  (~9.79\%)}& \num{83814} (~8.78\%)\\
\bottomrule
\end{tabular}
\end{table}

\begin{figure}[h!]
    \refstepcounter{suppfigure}
    \centering
    \includegraphics[width=0.7\linewidth]{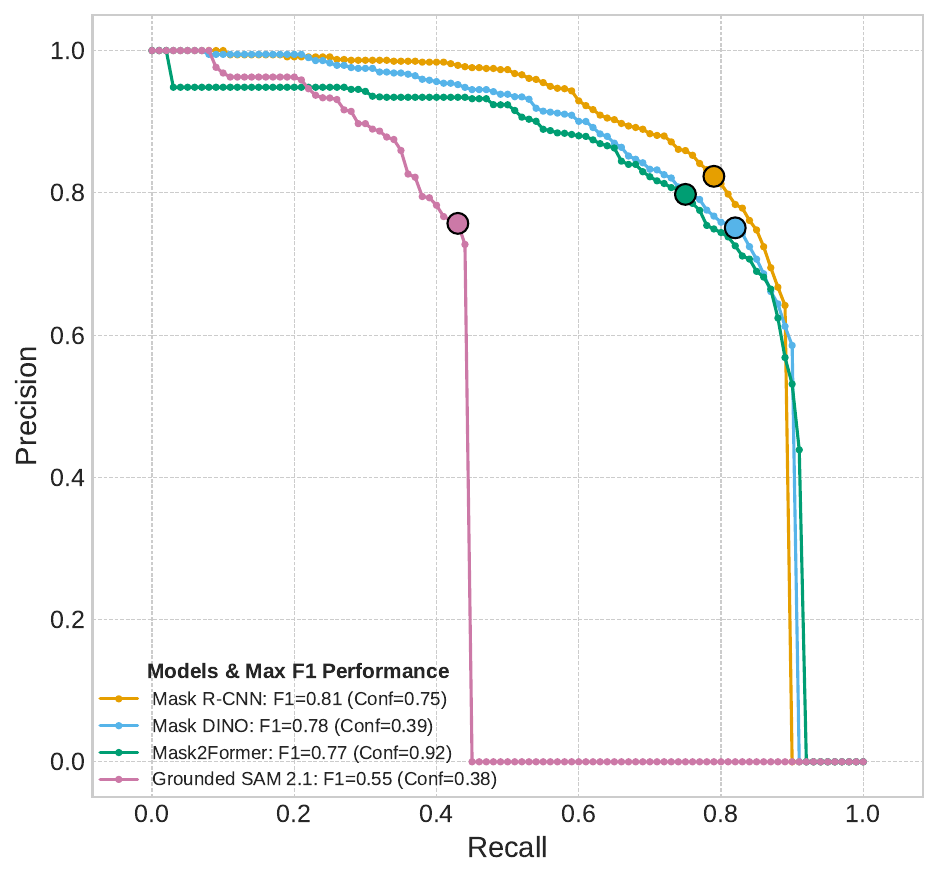}
    \caption{Precision-recall (PR) curves for the strongest zero-shot model (Grounded SAM 2.1) and the three supervised models (Mask R-CNN, Mask DINO, Mask2Former). We selected the confidence threshold for each model by finding the point on the PR curve with the highest F1-score.}
    \label{fig:pr}
\end{figure}

}
\end{document}